\documentclass[final,3p,times,authoryear]{elsarticle}

\usepackage{amssymb,amsmath}
\usepackage{algorithm,algorithmic}
\usepackage{array,booktabs,caption,subcaption,graphicx,verbatim,url,float,stfloats,textcomp,multirow,makecell,tabularx,geometry,pifont}
\usepackage{dsfont}
\usepackage[colorlinks=true, linkcolor=blue]{hyperref}
\usepackage{cleveref}
\usepackage[utf8]{inputenc}

\crefname{figure}{Fig.}{Figs.}
\newcommand{\figref}[1]{\hyperref[fig:#1]{Fig.~\ref*{fig:#1}}}
\crefname{table}{Table}{Tables}
\newcommand{\tabref}[1]{\hyperref[tab:#1]{Table~\ref*{tab:#1}}}
\newcommand{\lat}{\phi}
\newcommand{\lon}{\lambda}
\newcommand{\simi}[2]{\operatorname{sim}({#1}, {#2})}

\newcommand{\method}{BRIDGE-LC} % replace with \cfd{BRIDGE-LC} if color markup needed

\begin{document}

\begin{frontmatter}

\title{Geographical Context Matters: Bridging Fine and Coarse Spatial Information to Enhance Continental Land Cover Mapping}
           
\author[1]{Babak Ghassemi\corref{cor1}}
\author[2,3]{Cassio Fraga-Dantas}
\author[2,3]{Raffaele Gaetano}
\author[2,3]{Dino Ienco\corref{cor2}}
\author[1]{Omid Ghorbanzadeh}
\author[1]{Emma Izquierdo-Verdiguier}
\author[1]{Francesco Vuolo}

\address[1]{University of Natural Resources and Life Sciences, Vienna, Department of Ecosystem Management, Climate and Biodiversity, Institute of Geomatics, Peter Jordan Str. 82, Vienna, 1190, Austria}
\address[2]{TETIS, University of Montpellier, AgroParisTech, CIRAD/CNRS/INRAE, Montpellier, 34093, France}
\address[3]{INRIA, Université de Montpellier, Antenne INRIA de l’Université de Montpellier, Montpellier, 34090, France}

\cortext[cor1]{Corresponding author. Email: babak.ghassemi@boku.ac.at}
\cortext[cor2]{Corresponding author. Email: dino.ienco@inrae.fr}

\begin{abstract}

Land use and land cover mapping from Earth Observation (EO) data is a critical tool for sustainable land and resource management as, for instance, in domains like biodiversity and agricultural food production. While advanced machine learning and deep learning algorithms excel at analyzing EO imagery data, they often overlook crucial geospatial metadata information that could enhance scalability and accuracy across regional, continental, and global scales.
To address this limitation, we propose \method{} ({Bi-level Representation Integration for Disentangled GEospatial Land Cover}), a novel deep learning framework that explicitly integrates multi-scale geospatial information into the land cover classification process. By simultaneously leveraging fine-grained (latitude/longitude) and coarse-grained (biogeographical region) spatial information, our lightweight multi-layer perceptron architecture learns from both multi-scale information during training but only requires fine-grained information for inference, allowing it to disentangle region-specific from region-agnostic land cover features while maintaining computational efficiency comparable to standard machine learning approaches. 
To assess the quality of our framework, we use an open-access in-situ dataset spanning the 27 countries of the European Union and we adopt several competing classification approaches commonly considered for large-scale land cover mapping. We evaluated all the approaches through two scenarios: an extrapolation scenario in which training data encompasses samples coming from all the biogeographical regions and a leave-one-region-out scenario where samples from all the regions, except one, are employed for the training stage. Additionally, we also explore the spatial representation learned by the proposed deep learning model, highlighting a connection between its internal manifold and the geographical information used during the training stage. Our results demonstrate that integrating geospatial information improves land cover mapping performances, with the most substantial gains achieved by jointly leveraging both fine-grained and coarse-grained spatial information.
\end{abstract}

%%Graphical abstract
%\begin{graphicalabstract}
%\includegraphics{grabs}
%\end{graphicalabstract}

%%Research highlights
%\begin{highlights}
%\item Research highlight 1
%\item Research highlight 2
%\end{highlights}

%% Keywords

\end{frontmatter}

\section{Introduction}
\label{sec:intro}
Land Use and Land Cover (LULC) maps derived from Earth Observation (EO) data are essential information for sustainable land and resource management. They provide insights into how landscapes change over time due to human activity and natural processes. Predictions indicate that nearly ten billion people will reside on this planet by the year 2050. 
Thus, effectively monitoring farming landscapes and crop yield becomes crucial \citep{Bose2016Spiking, Luo2013Crop}. Precise and timely updated LULC maps play an important role here, supporting policy implementation and control and assisting in managing the spatial dynamics of agricultural food production \citep{rajadel2024within}.
  
Recent advancements in EO technology enable to observe Earth’s surface with unprecedented spatial, spectral, and temporal resolutions, facilitating the detection of subtle land changes across large areas~\citep{Cheng2023Application}. hese data characteristics not only benefits in mapping crop types and yields but it also provides significant value in disaster management, urban planning, and environmental protection. As a result, LULC maps are essential variables for both global and local monitoring and decision-making \citep{Zhao2023Land}.
The combination of open-access EO data and advanced machine learning (ML) techniques has enabled the development of precise, large-scale LULC maps~\cite {Gomes2020Overview}. The computational technology advancements of recent years have enabled the creation of numerous LULC products on a continental or global scale with spatial resolutions ranging from 10 meters to 1 kilometer. Satellite data sources used to derive these products include Sentinel, Landsat, MODIS, and AVHRR~\citep{ghassemi2024european}, among others.

Developments in ML and deep learning (DL) have enhanced LULC classification, especially by processing large amounts of EO data. Traditional ML algorithms, such as Random Forests (RF), Support Vector Machines (SVM), and K-Nearest Neighbors (KNN), are commonly adopted for large scale LULC mapping \citep{rajadel2024within,Amini2022Urban, GISLASON2006Random, Huang2002Assessment, Yisa2023Application, Mustafa2021Machine}. These methods significantly ameliorate classification accuracy and decrease processing time compared to manual interpretation \citep{Zhao2023Land}.
Recently, deep learning approaches have further advanced the way EO data are analyzed and interpreted, uncovering patterns and trends that improve the accuracy and quality of LULC classification \citep{Xie2022Deep}.
Multilayer Perceptrons~\citep{Sawada2020Monitoring, Zhang2021Downscaling} (MLP), Convolutional Neural Networks~\citep{Du2019Smallholder,Saralioglu2022Semantic} (CNN), Recurrent Neural Networks~\citep{SCHWALBERT2020Satellite, Jiang2020Deep} (RNN), Long short-term memory~\citep{campos2020understanding} (LSTM) and, more recently Transformers~\citep{garnot2020satellite}, are among the leading DL architectures that are currently exploited to derive effective LULC maps. However, despite the high-quality performance of these approaches in analyzing and understanding EO imagery, they generally neglect geospatial metadata information, such as pixel coordinates, which represent a distinct dimension inherent to EO data~\citep{rolf2024mission}. The integration of geospatial metadata is essential for enhancing the scalability and accuracy of EO-based analyses at regional~\citep{bellet2023land}, continental~\citep{russwurm2023geographic}, and global scales~\citep{mai2023sphere2vec}. Despite its importance, this information remains underutilized in tasks requiring geospatial awareness, such as large-scale land cover mapping. 

Recent studies are starting to explore how geospatial metadata can be utilized alongside imagery data to enhance tasks where geospatial knowledge is beneficial. Examples include species distribution modeling~\citep{russwurm2023geographic}, image super-resolution~\citep{panangian2025location} and scene classification~\citep{mai2023sphere2vec,bourcier2025learning}, proposing strategies for encoding and take advantages of this information.

While most recent approaches \citep{russwurm2023geographic,mai2023sphere2vec,bourcier2025learning} have focused on using geospatial metadata to pre-train DL models independently of the specific downstream task, only~\citep{Bellet2024} investigates the use of geographical coordinates as inputs for land cover mapping at a sub-country scale. Their results reveal that incorporating geospatial metadata enhances mapping accuracy.

To advance the integration of geospatial metadata into DL frameworks, specifically for large-scale land cover mapping, we propose a novel framework, namely \method{} ({Bi-level Representation Integration for Disentangled GEospatial Land Cover}), that explicitly incorporates both fine-grained and coarse-grained spatial information. This enhancement allows the DL model to take advantages from the geospatial information associated with the input data. Specifically, the proposed model integrates fine-grained information, such as geographic coordinates (latitude/longitude), through a learnable positional encoding. Additionally, it employs a biogeographical region partition\footnote{\url{https://www.eea.europa.eu/en/analysis/maps-and-charts/biogeographical-regions-in-europe-2}} (Alpine, Atlantic, BlackSea, Boreal, Continental, Mediterranean, Pannonian, and Steppic) to include coarse-level spatial information, aiming to differentiate between shared and specific geographical patterns. The backbone of our framework is a multilayer perceptron architecture, which offers two key advantages: (i) it provides a lightweight model with computational requirements comparable to standard ML approaches commonly employed for large-scale analysis, such as RF, and (ii) it allows for an unbiased assessment of multi-scale spatial information management, free from the complexities and potential biases of over-parametrized models. During training, \method{} leverages both fine-grained (latitude/longitude coordinates) and coarse-grained (biogeographical regions) spatial information. However, at inference time, the model only requires coordinate data for deployment. The biogeographical information guides the model during training to separate region-specific features from the common discriminative characteristics that describe the land cover classes. 

To evaluate the quality of our proposed framework, we utilize the publicly available analysis-ready data provided by~\citep{ghassemi2024european, d'Andrimont_LUCAS2022}, which encompasses in-situ samples and EO data from the 27 European Union countries (EU-27) \citep{lucas_copernicus_2022}. This benchmark serves as an ideal case study to test the capability of a land cover mapping framework to perform large-scale spatial inference. We consider two different scenarios: a first one where the training set contains data from all considered regions (referred to as the \emph{Extrapolation} scenario), and a second one where data from all regions except one are used for training, and the classification model is then tested on the held-out region (referred to as the \emph{Leave-one-region-out} scenario). We compare the performance of our framework against standard ML approaches, such as RF, SVM, and XGBoost, due to their widespread use in large-scale land cover mapping tasks. Additionally, we provide an in-depth analysis of the different components on which our framework is built on and we visually explore the location embeddings that are learnt by \method{}. Our results clearly demonstrate that integrating either fine-grained or coarse-grained geospatial information enhances performance. Moreover, the best results are achieved by jointly leveraging both fine-grained and coarse-grained spatial information, underscoring the effectiveness of our choice to explicitly integrate multi-scale geospatial information in the supervised learning process. 

The manuscript is organized as follows: study area and the associated data are described in Section~\ref{sec:data}. Section~\ref{sec:method} introduces the proposed framework, expanding on how coarse-grained and fine-grained geospatial information are integrated in the learning process. The experimental evaluation and the related findings are reported and discussed in Section~\ref{sec:results}, while Section~\ref{sec:conclu} draws the conclusions.

\section{Study area and data}
\label{sec:data}
\subsubsection{European Land Use and Cover Area Survey (LUCAS) 2022 Data}

Since 2006, the European Land Use and Cover Area frame Survey (LUCAS) has been conducted every three years, with surveys focused on LULC across the European Union. There have been $1,351,293$ observations at $651,780$ unique sites, encompassing $106$ variables, supported with $5.4$ million landscape photos from five editions of the LUCAS surveys \citep{d2020harmonised}.

In 2018, the scope of the survey was expanded with the addition of the "Copernicus module," which included $58,462$ polygonal geometries representing uniform land cover patches, each approximately $0.5$ hectares in size. These polygons provided comprehensive land cover (LC) and land use (LU) information, with $66$ LC categories (mostly vegetation cover) and $38$ LU classes \citep{d2020harmonised}. By 2022, the number of polygons had increased to $134,684$, accounting for $90.1\%$ of the original $149,429$ points from the Copernicus module, which is distributed across the EU-27. These polygons offer $75$ LC and $40$ LU classes, and their average size is around $0.2$ hectares following a new protocol \citep{lucas_copernicus_2022}. This development aimed to enhance data extraction, particularly for satellite images from Sentinel-1 (S1) and Sentinel-2 (S2).

\subsubsection{Classification Scheme Based on LUCAS 2022 Data}
\begin{figure}[h!]
    \centering
    \includegraphics[width=0.9\textwidth]{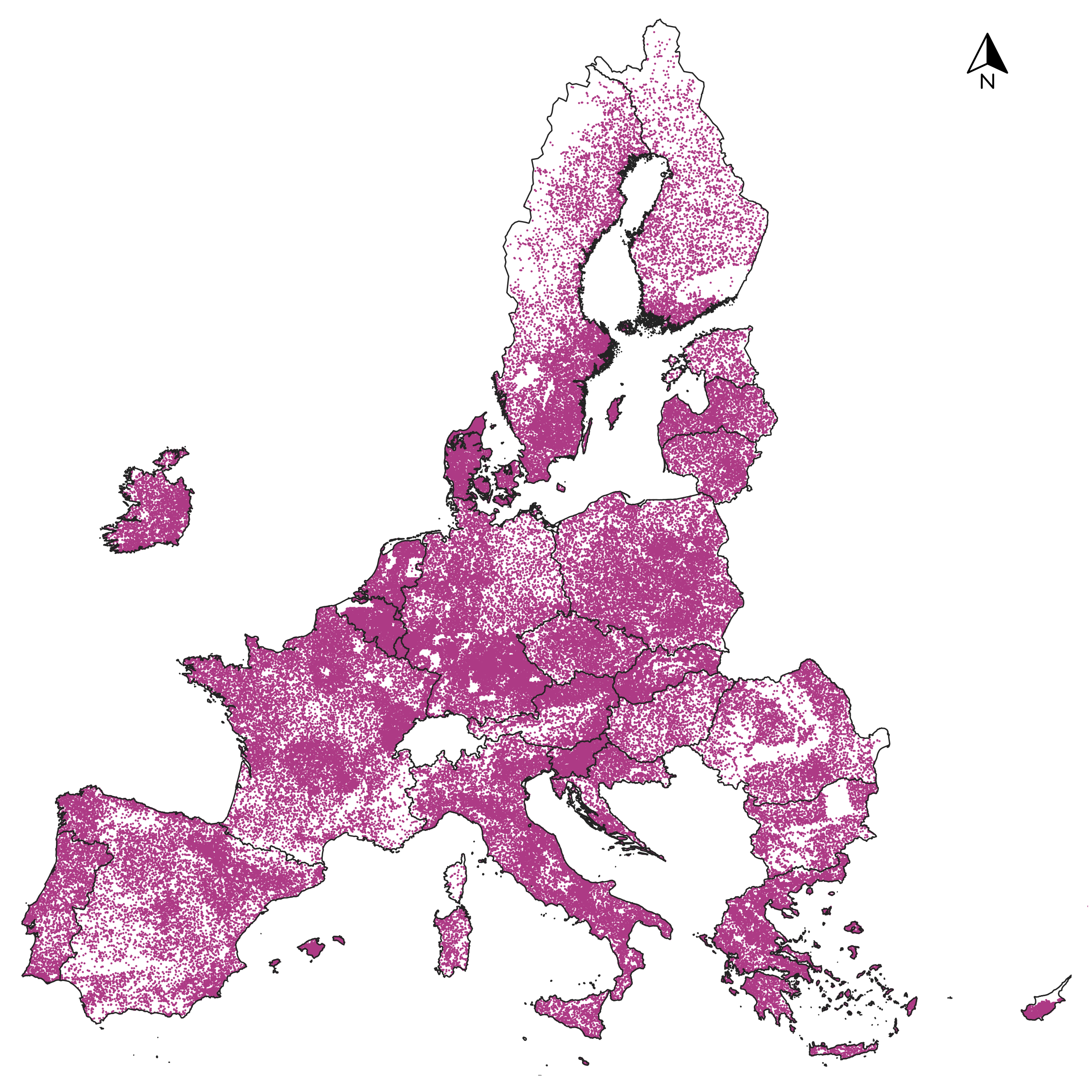}
    \caption{Database samples, distributed over the EU-27 countries.}
    \label{fig:samples}
\end{figure}

For our study, we use the publicly available database proposed by~\citep{ghassemi2024european} spanning all the EU-27. This public available database is based on the 2022 LUCAS data; it comprises seven primary LC categories at Level-1: 0-\textit{Woodland/Shrubland}, 1-\textit{Grassland}, 2-\textit{Bare land and lichens/moss}, 3-\textit{Wetlands}, 4-\textit{Artificial land}, 5-\textit{Water} and 7-\textit{Cropland} where the original LUCAS vegetation classes (Cropland, Woodland, Shrubland and Grassland) were re-organized into three main categories: Cropland, Woodland/Shrubland, and Grassland.
Furthermore, the 7-\textit{Cropland} category contains $19$ classes, each representing different crops or crop groups. Such a categorization is referred to as Level 2. The distribution of used samples across EU-27 can be found in \Cref{fig:samples}.

Table~\ref{tab:level1-classes} and Table~\ref{tab:level2-classes} report the Level-1 and Level-2 nomenclatures with the corresponding percentage proportions in terms of per-class samples, respectively.

\begin{table*}[!b]
\normalsize
\caption{~Level-1 classes \label{tab:level1-classes} and proportions.}
\centering
\begin{tabular}{c|c|c|l}
\hline
Class ID & $\#$ Samples & Proportion & Name \\ \hline
0 & 39,239 & $28.2\%$ & \textit{Woodland/Shrubland} \\ 
1 & 27,287 & $19.6\%$ & \textit{Grassland} \\ 
2 & 4,318 & $3.1\%$ & \textit{Bare land and lichens/moss} \\ 
3 & 4,539 & $3.3\%$ & \textit{Wetlands} \\ 
4 & 14,122 & $10.1\%$ & \textit{Artificial land} \\ 
5 & 4,104 & $2.9\%$ & \textit{Water} \\ 
6 & 45,608 & $32.8\%$ & \textit{Cropland} \\ \hline \hline
Total $\#$ Samples & 139,217 & $100\%$ & 
\end{tabular}
\end{table*}

\begin{table*}[!ht]
\caption{~Level-2 classes and proportions \label{tab:level2-classes}. These sub-classes are part of the Level-1 class Cropland.}
\centering
\begin{tabular}{c|c|c|l}
\hline
Class ID & $\#$ Samples & Proportion & Name \\ \hline
0  & 2,857   & $8.3\%$    & Bare arable land \\ 
1  & 8,073   & $23.3\%$ & Common wheat \\ 
2  & 1,785   & $2.6\%$  & Durum wheat \\ 
3  & 3,866   & $11.2\%$ & Barley \\ 
4  & 1,915   & $2.9\%$  & Rye \\ 
5  & 1,974   & $3.2\%$  & Oats \\ 
6  & 5,262   & $15.2\%$ & Maize \\ 
7  & 1,809   & $2.0\%$  & Potatoes \\ 
8  & 1,829   & $2.1\%$  & Sugar beet \\ 
9  & 980        & $0.7\%$  & Other roots crops \\ 
10 & 2,514   & $7.3\%$  & Rape and turnip rape \\ 
11 & 1,786   & $2.0\%$  & Other non-permanent industrial crops \\ 
12 & 1,915   & $3.1\%$  & Dry pulses, vegetables and flowers \\ 
13 & 2,602   & $7.5\%$  & Other fodder crops \\ 
14 & 858      & $0.6\%$  & Other cereals \\ 
15 & 1,963   & $4.0\%$  & Sunflower \\ 
16 & 1,642   & $1.2\%$  & Soya \\ 
17 & 1,900   & $2.7\%$  & Triticale \\ 
18 & 78        & $0.1\%$  & Rice \\ \hline \hline
Total $\#$ Samples &  45,608 & $100\%$ &
\end{tabular}
\end{table*}

\subsection{Earth Observation Data}

Various EO information were used to describe the study regions. This information includes high-resolution, frequent-revisit reflectance data from S2 and backscatter coefficients from S1, along with auxiliary information like Land Surface Temperature (LST) and Digital Elevation Model (DEM) data. The following sections describe, in more detail, these data sources and the features utilized.

\subsubsection{Sentinel-2 Data}
The S2 mission, developed by the EU within the framework of the Copernicus program, comprises two identical satellites, Sentinel-2A %\emma{}{(currently replaced by Sentinel-2C)} 
and Sentinel-2B to gather spatio-temporal high-resolution optical images of the Earth. These satellites collect data in $13$ spectral bands covering the visible to the infrared spectrum, with spatial resolutions ranging from $10$ to $60$ meters. This broad spectral range, along with global coverage and a five-day revisit cycle, enable in-depth investigation of vegetation, land cover, water bodies, and other features \citep{drusch2012sentinel}.\\

S2 Level-2A products from Google Earth Engine (GEE) were used. These products offer surface reflectance values that are corrected from atmospheric effects. The "harmonized" aspect of the collection compensates for reflectance offsets implemented by the European Space Agency on January $21^{st}$, 2022. The Scene Classification (SCL) band in S2 data indicates per-pixel values for cloud cover, shadows, and other anomalies. The study used S2 images from January 1 to December $31^{st}$, 2022, with cloud fraction cover below $50\%$. Pixels with cloud probabilities above $75\%$ or flagged as Saturated or Defective, Clouds High Probability, Cirrus, and Snow/Ice in the SCL information were excluded. These thresholds were determined using a visual, trial-and-error approach. To ensure consistent spatial resolution, $20 m$ bands were resampled to $10 m$ using the nearest neighbor method.

A range of spectral bands (B02-B08, B8A, B11, and B12) and $15$ spectral indices, and a biophysical parameter, Leaf Area Index (LAI), were employed in the study. Among $15$ spectral indices, five are focused on vegetation: Enhanced Vegetation Index 2 (EVI2) \citep{jiang2008development}, Green Normalized Difference Vegetation Index (GNDVI) \citep{gitelson1996use}, Leaf Area Index green (LAIg) \citep{pasqualotto2019multi}, Leaf Chlorophyll Content Index (LCCI) \citep{wulf2015sentinel} and Normalized Difference Vegetation Index (NDVI) \citep{kriegler1969preprocessing}. Four of them are concentrated on Soil characteristics: Bare Soil Index (BSI) \citep{rikimaru2002tropical}, Modified Soil Adjusted Vegetation Index (MSAVI) \citep{qi1994modified}, Normalized Difference Tillage Index (NDTI) \citep{van1997using}, and Soil Adjusted Vegetation Index (SAVI) \citep{huete1988soil}. Two are urban indices: Built-up Land Features Extraction Index (BLFEI) \citep{bouhennache2019new} and Normalized Difference Built-up Index (NDBI) \citep{zha2003use} and two, water indices: Modified Normalized Difference Water Index (MNDWI) \citep{xu2006modification} and Normalized Difference Water Index (NDWI) \citep{mcfeeters1996use}. Besides, the difference between Red and Shortwave Infrared 1 bands (DIRESWIR) \citep{jacques2014monitoring} and the ratio between Near Infrared and Red bands (SRNIRR) \citep{blackburn1998quantifying}; both were also used. \tabref{s2features} reports detailed information about the used indices. 

As a result of cloud cover at specific regions, no data values appear in the S2 feature datasets. Therefore, monthly and annual temporal composite products were created. In this study, only the annual $5\textsuperscript{th}$, $50\textsuperscript{th}$, and $98\textsuperscript{th}$ percentiles for each band and index were used resulting in $78$ features ($26$ × $3\textsubscript{yearly percentile}$).

\begin{table}[h!]
    \centering
    \caption{~Spectral bands, indices, and a biophysical parameter extracted from S2 [Notation: NIR = Near Infrared; SWIR = Shortwave Infrared; WL = Wavelength].}
    \label{tab:s2features}
    \begin{tabular}{@{}p{5cm}@{\hskip 5pt}p{8.5cm}@{}}
        \hline
        \textbf{Feature Name} & \textbf{Description} \\ 
        \hline
        \multirow{10}{5cm}{\centering \textbf{Spectral Bands}} & \textbf{B2:} Blue (WL: 496.6 nm (S2A) / 492.1 nm (S2B)) \\ 
        & \textbf{B3:} Green (WL: 560 nm (S2A) / 559 nm (S2B)) \\ 
        & \textbf{B4:} Red (WL: 664.5 nm (S2A) / 665 nm (S2B)) \\ 
        & \textbf{B5:} Red Edge 1 (WL: 703.9 nm (S2A) / 703.8 nm (S2B)) \\ 
        & \textbf{B6:} Red Edge 2 (WL: 740.2 nm (S2A) / 739.1 nm (S2B)) \\ 
        & \textbf{B7:} Red Edge 3 (WL: 782.5 nm (S2A) / 779.7 nm (S2B)) \\ 
        & \textbf{B8:} NIR (WL: 835.1 nm (S2A) / 833 nm (S2B)) \\ 
        & \textbf{B8A:} NIR narrow (WL: 864.8 nm (S2A) / 864 nm (S2B)) \\ 
        & \textbf{B11:} SWIR 1 (WL: 1613.7 nm (S2A) / 1610.4 nm (S2B)) \\ 
        & \textbf{B12:} SWIR 2 (WL: 2202.4 nm (S2A) / 2185.7 nm (S2B)) \\ 
        \hline
        \multirow{16}{5cm}{\centering \textbf{Spectral Indices and} \\ \textbf{Biophysical Parameters}} & \textbf{BLEFI:} $(((B3 + B4 + B12)/3) - B11) / (((B3 + B4 + B12)/3) + B11)$ \\ 
        & \textbf{BSI:} $((B11 + B4) - (B8 + B2)) / ((B11 + B4) + (B8 + B2))$ \\ 
        & \textbf{DIRESWIR:} $B4 - B11$ \\ 
        & \textbf{EVI2:} $(B8 - B4) / (B8 + B4 + 1) * 2.4$ \\ 
        & \textbf{GNDVI:} $(B8 - B3) / (B8 + B3)$ \\ 
        & \textbf{LAI:} $3.618 * (2.5 * ((B8 - B4) / (B8 + 6 * B4 - 7.5 * B2 + 1))) - 0.118$ \\ 
        & \textbf{LAIg:} $5.405 * ((B8A - B5) / (B8A + B5)) - 0.114$ \\ 
        & \textbf{LCCI:} $B7 / B5$ \\ 
        & \textbf{MNDWI:} $(B3 - B11) / (B3 + B11)$ \\ 
        & \textbf{MSAVI:} $0.5 * (2 * B8 + 1 - sqrt((2 * B8 + 1)^2 - 8 * (B8 - B4)))$ \\ 
        & \textbf{NDBI:} $(B11 - B8) / (B11 + B8)$ \\ 
        & \textbf{NDTI:} $(B11 - B12) / (B11 + B12)$ \\ 
        & \textbf{NDVI:} $(B8 - B4) / (B8 + B4)$ \\ 
        & \textbf{NDWI:} $(B8 - B11) / (B8 + B11)$ \\ 
        & \textbf{SAVI:} $(B8 - B4) / (B8 + B4 + 0.5) * 1.5$ \\ 
        & \textbf{SRNIRR:} $B8 / B4$ \\ 
        \hline
    \end{tabular}
\end{table}

\subsubsection{Sentinel-1 Data}

The S1 mission, part of the European Union's Copernicus program, operates with two synthetic aperture radar (SAR) satellites, Sentinel-1A (S1A) and Sentinel-1B (S1B). These satellites provide extensive global coverage, with a revisit time of approximately $6$ days. However, after the failure of S1B in December 2021, the revisit interval increased to $12$ days till December 2024, with the start of operations of the Sentinel-1C (S1C) satellite. S1 operates in the C-band ($5.4$ GHz) and employs SAR, which transmits vertically polarized microwave signals to the Earth's surface and records backscattered signals in both vertical (VV) and horizontal (VH) polarizations. This radar system benefits agricultural applications because it can penetrate clouds and capture data regardless of weather or lighting conditions, day or night. Crop characteristics such as leaf density, canopy moisture content, and soil surface properties like moisture levels and roughness all impact radar backscatter measurements~\citep{torres2012gmes}.

Over Europe, S1 collects data in both ascending and descending orbits. The Interferometric Wide (IW) mode, with a $10 m$ sampling distance, is the standard mode used for global land. While S1 provides freely accessible data, it is available in Level-1 formats (Ground Range Detected (GRD) and Single Look Complex (SLC)), which require further processing. In GEE, the GRD scenes were pre-processed using a predefined workflow from the S1 SNAP Toolbox~\footnote{\url{http://step.esa.int}}, which includes geocoding with the Shuttle Radar Topography Mission (SRTM) $90 m$ DEM. The obtained sigma naught backscatter coefficients (\(\sigma_0\)) are available in the COPERNICUS/S1\_GRD\_FLOAT collection. Both VV and VH bands are used for IW mode.

A speckle filter based on the focal median method was applied to reduce noise in the radar imagery, utilizing a circular kernel with a $30 m$ radius \citep{kupidura2016comparison}. The monthly median and the yearly $5\textsuperscript{th}$, $50\textsuperscript{th}$, and $98\textsuperscript{th}$ percentile values for VV and VH bands and various indices were computed (\tabref{s1features}). These indices, including VV/VH, VV-VH, Ratio Vegetation Index (RVI)\citep{arii2010general}, Normalized Difference Polarization Index (NDPI) \citep{mitchard2012mapping}, and modified Dual Polarimetric SAR Vegetation Index (DPSVIm) \citep{dos2021vegetation}, help enhance LULC classification. As microwave data is unaffected by cloud cover, monthly median values for all months of the year were included in the analysis. In total, $105$ features ($7$ × ($12\textsubscript{monthly median}$ + $3\textsubscript{yearly percentile}$)) were derived from the S1 data for classification, covering both monthly and yearly data. 

\begin{table}[h!]
    \centering
    \caption{~Microwave bands and indices obtained from S1 }
    \label{tab:s1features}
    \resizebox{\columnwidth}{!}{% Adjusting to fit the column width
    \begin{tabular}{@{}p{5cm}@{\hskip 5pt}p{10cm}@{}} % Adjusting column widths
        \hline
        \textbf{Feature Name} & \textbf{Description} \\ 
        \hline
        \multirow{7}{*}{\centering \textbf{Microwave features}} 
        & \textbf{VV:} Single co-polarization, vertical transmit/vertical receive \\ 
        & \textbf{VH:} Dual-band cross-polarization, vertical transmit/horizontal receive \\ 
        & \textbf{VV/VH:} The ratio between the VV and the VH polarizations \\ 
        & \textbf{VV-VH:} The difference between the VV and the VH polarizations \\ 
        & \textbf{RVI:} \(\frac{4\sigma_{VH}^0}{\sigma_{VV}^0+\sigma_{VH}^0}\) \\ 
        & \textbf{NDPI:} \(\frac{\sigma_{VV}^0 - \sigma_{VH}^0}{\sigma_{VV}^0+\sigma_{VH}^0}\) \\ 
        & \textbf{DPSVIm:} \(\frac{\sigma_{VV}^0 \sigma_{VV}^0 + \sigma_{VV}^0 \sigma_{VH}^0}{\sqrt{2}}\) \\ 
        \hline
    \end{tabular}%
    }
\end{table}

\subsubsection{Auxiliary temperature and elevation data}

Along with S1 and S2 data, auxiliary surface temperature and elevation features that could affect LULC classification were used. Land Surface Temperature (LST) and elevation data were utilized to supplement the spectral and microwave datasets. 

The study took advantage of the LST\_Day feature of the MOD21C3 product from the MODIS (Moderate Resolution Imaging Spectroradiometer) sensor installed on the Terra and Aqua satellites~\citep{mod21c3.061}. This information provides monthly composite LST data at a resolution of $1 km$, making it ideal for capturing temperature patterns over time. 

The Global Digital Elevation Model (GDEM) v3, obtained from the ASTER instrument on the Terra satellite, was employed to account for elevation~\citep{astgtm.003}. With $30 m$ resolution, the GDEM provides a detailed terrain model that captures variations in height and shape that affect LULC classification. 

All auxiliary data were resampled to match the main data's spatial resolution of 10 m using the nearest neighbor approach, guaranteeing uniformity across all data layers. This investigation included $13$ auxiliary features, comprising $12$ monthly features for LST data and one for DEM data.

\subsubsection{Fine and Coarse grained geographical information}

In addition to the aforementioned auxiliary data, the latitude and longitude of samples were employed as input information for positional encoding, which enable the model to incorporate spatial context by transforming raw coordinates into a representation that captures relative geographic relationships. Furthermore, biogeographical region information provided by European Environment Agency (EEA)%
\footnote{\url{https://www.eea.europa.eu/data-and-maps/figures/biogeographical-regions-in-europe-2}}, comprising $8$ unique values (Alpine, Atlantic, Black Sea, Boreal, Continental, Mediterranean, Pannonian, and Steppic) in the study area, were utilized as supervision signals on the proposed %disentanglement 
approach. 

\begin{figure}
    \centering
    \includegraphics[width=0.9\textwidth]{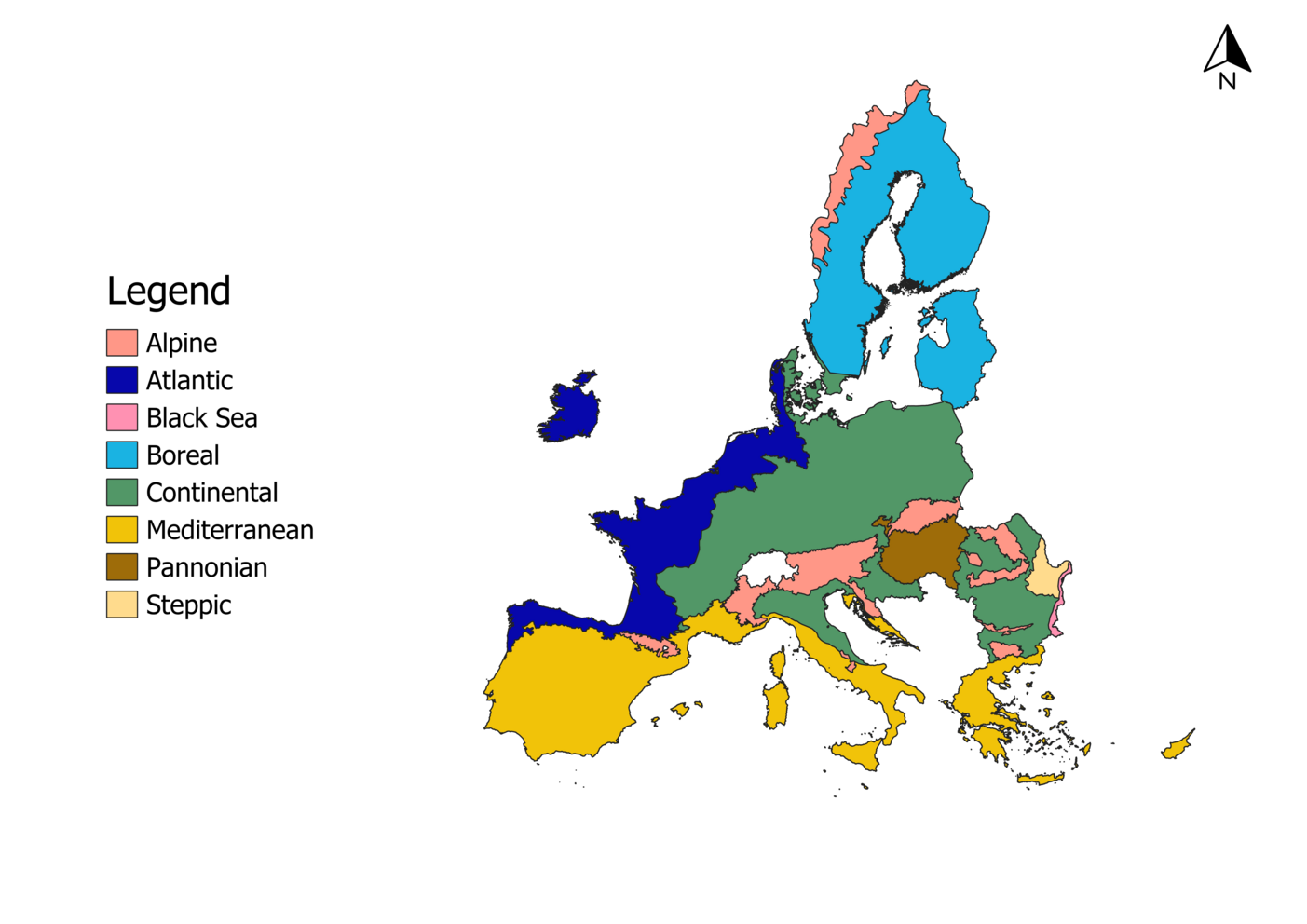}
    \caption{Subset of the biogeographical regions produced by the European Environment Agency (EEA).}
    \label{fig:regions}
\end{figure}

The distribution of the available labelled data in terms of the climate regions is shown in \Cref{tab:regions_distrib}. 
% NaN samples
A total of $43$ samples with uninformed climatic region where removed from the dataset.
\begin{table*}[b!]
\normalsize
\caption{~Number of samples per climate region\label{tab:regions_distrib}}
\centering
\begin{tabular}{c|cccc|}
Climate Reg	 & Alpine & Atlantic &	BlackSea &	Boreal \\ 
	 \hline
& \multicolumn{4}{c|}{\textbf{Level 1}} \\
\hline
$\#$ samples & 6615 & 21587 & 258 & 19109 \\
Percentage & 4.8\%	& 15.5\% &	0.2\% &	13.7\%  \\ \hline
& \multicolumn{4}{c|}{\textbf{Level 2}} \\
\hline
$\#$ samples & 812	& 9421	& 128	& 2612		\\
Percentage & 1.8\%	 &	20.7\%	& 0.3\%		& 5.7\%		 \\ \hline
\hline
Climate Reg &	Continental &	Mediterranean &	Pannonian &	Steppic \\
\hline
& \multicolumn{4}{c|}{\textbf{Level 1}} \\
\hline
$\#$ samples & 54553 & 31509 & 4749 & 794\\
Percentage &	39.2\% &	22.6\% &	3.4\% &	0.6\% \\
\hline
& \multicolumn{4}{c|}{\textbf{Level 2}} \\
\hline
$\#$ samples & 23213 & 	7348 & 1653	& 419\\
Percentage & 50.9\% & 16.1\% & 3.6\%	& 0.9\% \\
\hline
\end{tabular}
\end{table*}

\subsection{Dataset preparation and balancing}

From the resulting $196$ features ($78$ from S2, $105$ from S1, and $13$ from auxiliary data), a subset of $109$ %effective 
features were selected for the classification procedures, following the analysis in \citet{ghassemi2024european} where these features were picked based on their prominence provided by a Random Forest classification model. 

To prepare the dataset for training and testing, feature values were extracted from the centroids of the $134,670$ available polygons in the LUCAS 2022 Copernicus dataset. Level-1 and Level-2 class labels were assigned according to the scheme provided in \citet{d2021parcel}. After removing polygons without corresponding LC labels and eliminating samples with missing LST data, the sample size was reduced to $133,813$. Among these samples, $81,427$ that were not affected by cloud cover issues were balanced based on their label counts. This was done utilizing Euclidean distance criteria and an RF classifier to reduce samples from overrepresented classes and increase samples for underrepresented classes. Similar data points were eliminated for classes with abundant samples, while sparse classes were supplemented with additional samples from within polygons, not limited to centroids \citep{ghassemi2022designing}. Consequently, after balancing, the number of samples rose from $81,427$ to $86,831$. More details about the dataset preparation procedure are available in \citet{ghassemi2024european}.
Finally, after applying the aforementioned pre-processing steps, $139,217$ labeled samples were made available for the classification procedure.

\section{Methodology}
\label{sec:method}
\begin{figure*}[t]
    \centering
    \includegraphics[width=\textwidth,trim={0 0 1.5cm 0},clip]{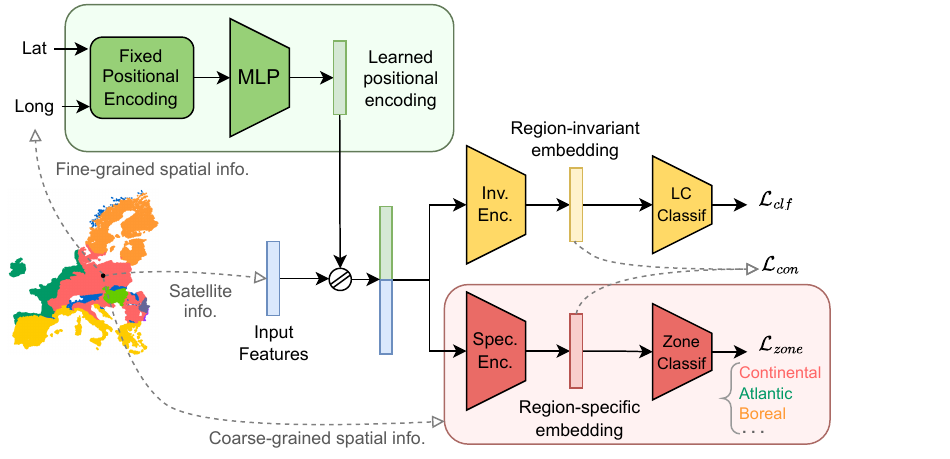}
    \caption{Proposed framework. In addition to the standard land cover classification branch (yellow), which processes reflectance-based input features, we introduce two complementary components: (1) a fine-grained spatial information branch (green) utilizing latitude-longitude coordinates, and (2) a coarse-grained spatial information branch (red) based on biogeographical region labels. During training, the red branch guides the model to learn region-invariant embeddings that are more robust and generalizable for land cover classification. This branch is used only during training and can be discarded at inference time.
    }
    \label{fig:method}
\end{figure*}

The proposed approach leverages coarse-grained (biogeographical regions), fine-grained spatial information (geographical coordinates), and spectral and temporal information to improve land cover classification performance. An overview of the proposed DL framework, detailed in this section, is given in \Cref{fig:method}.

\subsection{Overall structure of the new classification approach}

The proposed framework consists of three main components:
\begin{enumerate}
    \item The ``Positional encoding module" handles the fine-grained (Lat./Long.) spatial information.
    \item The ``Region-specific branch" handles the coarse-grained (biogeographical regions) information by generating region-specific embeddings that allow the discrimination of samples from the different biogeographical regions through a classifier.   
    \item The ``Land cover classification branch" utilizes both radiometric and positional information to classify land cover types. Through a feature disentanglement approach, it is designed to complement the Region-specific branch, ensuring that the generated embeddings capture discriminative information while remaining invariant to biogeographical provenance.
    
\end{enumerate}

These three components are active during the training phase, where the proposed network receives input data from diverse locations in Europe and learns to integrate the additional spatial information to improve the final land cover classification performance. The fine-grained spatial information (Latitude and Longitude) is encoded in the form of a learned positional encoding vector, as detailed in \Cref{ssec:pos-enc}, which is then concatenated to the original reflectance based input features. This location-enriched feature vector is the actual input of our land cover classification module as well as its parallel region-specific branch. Together, these two branches compose a \emph{feature disentanglement} module, detailed in \Cref{ssec:disentanglement}, which allows the proposed model to effectively utilize the coarse-grained biogeographical region information provided during training. This is achieved by learning how to isolate region-specific information that could otherwise hinder geographical generalization across the vast extent of the studied area. The training procedure is described in \Cref{ssec:training} and implementation details are provided in \Cref{ssec:arch_details}.

Once trained, the proposed model can be deployed for inference. At this stage, the region-specific branch is dropped, as only the region-invariant embedding is used for the downstream land cover classification task. At this point, the invariant encoder has already learned how to extract discriminant features that remain, to some extent, agnostic to the biogeographical region of provenance, thanks to the joint training procedure performed with the support of the region-specific branch.

\subsection{Fine-grained: Positional encoding} \label{ssec:pos-enc}

% Fixed positional encoding
Given a sample's geographical coordinates, latitude and longitude $(\lat,\lon)$ in degrees, a fixed positional encoding vector $\mathbf{p}_\text{enc} = g(\lat,\lon) \in \mathbb{R}^{2d}$ with embedding space dimension $2d$ is computed according to the equations below:

\begin{align}
g : \mathbb{R}^2 &\to \mathbb{R}^{2d} \nonumber\\
(\lat,\lon) &\mapsto \mathbf{p}_{\text{enc}} =
\begin{bmatrix} \mathbf{p}_\lat \\ \mathbf{p}_\lon \end{bmatrix} %\in \mathbb{R}^{2d} \quad \text{where for } ~ i \in \{0, \dots, d/2 - 1\} 
\label{eq:pos_enc1}
\end{align}
where the encodings of latitude and longitude $\mathbf{p}_\lat$ and $\mathbf{p}_\lon  \in \mathbb{R}^{d}$ are defined as follows, with $i \in \{0, \dots, d/2 - 1\} $ and $n = 10^4$:
%where, for $i \in \{0, \dots, \dfrac{d}{2} - 1\} $ and $n = 10000$, the encodings $\mathbf{p}_x, \mathbf{p}_y  \in \mathbb{R}^{d}$ are defined as follows
\begin{align}
\left\{
\begin{aligned}
&\mathbf{p}_\lat(2i) = \sin(n^{-2i/d}\lat)    \\
&\mathbf{p}_\lat(2i+1) = \cos(n^{-2i/d}\lat)  
\end{aligned}
\right.
,
\quad
\left\{
\begin{aligned}
&\mathbf{p}_\lon(2i) = \sin(n^{-2i/d}\lon)    \\
&\mathbf{p}_\lon(2i+1) = \cos(n^{-2i/d}\lon) 
\end{aligned}
%\text{ with }  0 \leq i < d/2 
\right. \label{eq:pos_enc2}
\end{align}

This positional encoding scheme is inspired by recent literature in Transformers models \citep{vaswani2017attention}, first proposed in Natural Language Processing (NLP) to provide information on a token's position on a text. More recently, this approach has been extended to the computer vision domain, where the $(x,y)$ position of a patch in the image is encoded. Here, similarly, 2-dimensional positional information is encoded by concatenating the respective 1-dimensional encodings. 
A similar approach has been used to encode geographical coordinates in \citep{Baudoux2021, Bellet2024}.

% Learned positional encoding
This closed-form positional encoding vector is then fed to a Multi-Layer Perceptron (MLP), $f_\theta: \mathbb{R}^{2d} \to \mathbb{R}^{2d}$, which outputs the final positional representation: 
\begin{equation}
  \mathbf{p}= f_\theta(\mathbf{p}_{\text{enc}} ). 
  \label{eq:pos_enc_MLP}
\end{equation}
This MLP module is learned jointly with the other components of the proposed architecture. It is, therefore, trained in an end-to-end manner to provide exploitable information related to the geographical location for the downstream classification task.

\subsection{Feature disentanglement}\label{ssec:disentanglement}

Data originating from diverse geographical regions often exhibit significant variations in distribution, inevitably presenting some degree of region-dependant intrinsic characteristics. Directly combining such data for deep learning model training without explicitly tackling such distribution shifts can lead to sub-optimal model performance. 

We propose a feature disentanglement approach inspired by recent domain adaptation literature \citep{Dantas2024,JoY23} to address these challenges. The idea is to separate the share of information specific to a given biogeographical region from generic class-discriminative features that remain consistent across geographical locations. By doing so, the classifier, which operates on a unified set of classes across the studied regions, is provided with a more coherent set of features for different samples from the same class, regardless of their biogeographical region of provenance. This not only makes the classifier's task easier (as it is trained on more homogeneous input features), but also enhances the model's overall generalization capabilities(as it becomes more robust to variations associated to different geographical locations).

To perform feature disentanglement, we utilize a dual-branch architecture, each encoding one type of information. Namely: 1) Region-invariant; 2) Region-specific branch. The branches, structured similarly, are composed of an encoder, which extracts specific or invariant features, followed by a classifier (cf. \Cref{fig:method}). For the region-invariant branch, the extracted features are used for the land cover classification task. The features extracted by the region-specific branch, in turn, are used for a region classification task, which helps enforce region-specific information to be channeled to this part of the architecture. Additionally, we employ a contrastive loss to enforce orthogonality between the encoded information in the two branches while also promoting a more structured representation space. 
As each branch is trained to encode complementary information, we expect the region-invariant one to primarily capture features that generalize across regions, ultimately resulting in a more robust land cover classification performance in a diverse range of biogeographical regions.
Further details on the specifics of the architecture (e.g., width and number of layers) are provided in \Cref{ssec:arch_details}.

\subsection{Training procedure} \label{ssec:training}

The training dataset $\{(\mathbf{x}_i, y_i)\}_{i=1}^{N_s}$ is composed of $N_s$ labelled input samples, where $\mathbf{x}_i \in \mathbb{R}^{F}$ denotes the set of $F$ input features associated to the $i$-th sample and $y_i \in \{1, \dots, C\}$ the corresponding label among the $C$ considered classes. Additionally, to every sample is associated with some accessory information related to its fine and coarse-grained geographic location: $\{(\lat, \lon)_i\}_{i=1}^N$ for the latitude-longitude coordinates and $\{ r_i\}_{i=1}^N$ for its biogeographical region label, 
%---where $m_i=1$ if the $i$-th sample belongs to the Mediterranean region and $m_i=0$ otherwise. 
with $r_i \in \{1, \dots, R\}$ the index of one of the $R = 8$ considered regions in the following order:
\[
\{\text{Alpine},\, \text{Atlantic},\, \text{Black Sea},\, \text{Boreal},\, \text{Continental},\, 
\text{Mediterranean},\, \text{Pannonian},\, \text{Steppic}\}
\]

The F-dimensional array of reflectance-based input features $\mathbf{x}_i$ is concatenated to the corresponding spatial positional encoding $\mathbf{p}_i$ to form the actual input of the classification modules $(\mathbf{x}\|\mathbf{p})_i \in \mathbb{R}^{F+2d}$.

This enriched input vector is given to both region-invariant and region-specific encoders $f_{inv}$, $f_{spec}: \mathbb{R}^{F+2d} \to \mathbb{R}^D$ which generate embedding vectors $\mathbf{z}_{inv}$ and $\mathbf{z}_{spec} \in \mathbb{R}^{D}$, respectively, with $\mathbf{z}_{inv}=f_{inv}(\mathbf{x}\|\mathbf{p})$ and $\mathbf{z}_{spec}=f_{spec}(\mathbf{x}\|\mathbf{p})$.
% Classification loss - task
Then, the region-invariant embedding is fed to a land-cover classifier module, which outputs the predicted class distribution vector $\mathbf{\hat{y}}_i \in \mathbb{R}^C$ on which we compute the following categorical cross-entropy loss, denoting $\mathbf{\hat{y}}_{i}(k)$ the predicted probability for class $k$ on the $i$-th sample:
\begin{equation}
\mathcal{L}_{LC} = \sum_{k=1}^C \mathds{1}(y_i = k) \log \left(\mathbf{\hat{y}}_{i}(k)\right)
\end{equation}
% Classification loss - ecoregion
The latter embedding $\mathbf{z}_{spec}$, in its turn, is fed to a separate classifier 

which tries to predict the  biogeographical region information of the given sample and is also assessed with a categorical cross entropy-loss, where we denote 
%$\hat{m}_i \in [0, 1]$ the predicted probability of belonging to the Mediterranean region:
%\begin{equation}
%    \mathcal{L}_{region} = m_i \log(\hat{m}_i) + (1 - m_i) \log(1 - %\hat{m}_i)
%\end{equation}
$\mathbf{\hat{r}}_i \in \mathbb{R}^R$ the predicted probability vector on each of the $R$ considered regions and $r_i$ contains the corresponding ground-truth information:
\begin{equation}
\mathcal{L}_{region} = \sum_{k=1}^R \mathds{1}(r_i = k) \log \left(\mathbf{\hat{r}}_{i}(k)\right)
\end{equation}

% Contrastive loss
To further decouple the two separate branches, we employ contrastive loss, which has shown promising results for feature disentanglement in previous works~\citep{Dantas2024}. 
More precisely, a supervised contrastive loss \citep{khosla2020supervised} is employed to better structure the extracted region-invariant and region-specific embeddings. 
In this loss,  the samples are split into some pre-defined sub-categories (classes), which are pulled together in the embedding space to form clusters within the same class while simultaneously pushing them away from all samples from different classes.

Here, we define a set $C + R$ classes for the supervised contrastive loss: $C$ for the region-invariant embeddings (according to its corresponding land-cover class) and $R$ for the region-specific embeddings, where samples from different biogeographical regions are mapped to separate categories. In particular, note that region-invariant embeddings from the same land-cover class but coming from different biogeographical regions are purposely mapped to the same class in the defined scheme. Similarly, region-specific embedding is never mapped to the same class as region-invariant embedding, which will contribute to keeping them dissimilar.

Consider that training is performed over batches of input data composed by $B$ samples each. Given a current batch $\{(\mathbf{x}_i, y_i)\}_{i=1}^{B}$, and the associated embeddings $\{(\mathbf{z}_{inv}, \mathbf{z}_{spec})_i\}_{i=1}^{B}$,
the final supervised contrastive loss is obtained through a series of pairwise comparisons between similar and dissimilar embeddings within the batch. More precisely, pairs of similar elements (i.e., those belonging to the same category among the $C+R$ described above) are referred to as \emph{positive} pairs and are further pulled together, while pairs of dissimilar elements (i.e., from different categories) are referenced as \emph{negative} pairs and are pushed apart.

The supervised contrastive loss is computed as a sum where each element of the batch, one at a time, is taken as the current \emph{anchor}---that is, the reference sample to be paired with all other elements in the batch. By iterating over all samples as anchors, all possible pairings are explored.
Since each input sample yields two separate embeddings, $\mathbf{z}_{inv}$ and $\mathbf{z}_{spec}$, a batch of size $B$ results in a total of $2B$ embeddings to be paired. 
Let us denote by $I := \{1, \dots, 2B\}$ the index set corresponding to these embeddings, and denote the $i$-th embedding simply as $\mathbf{z}_i$ for each $i \in I$.

The resulting supervised contrastive loss is defined as follows:
\begin{align} \label{eq:supervised_contrastive_loss}
    \mathcal{L}_{con} = -\!\sum_{i\in I} \frac{1}{|P(i)|} \sum_{p \in P(i)} \log \frac{\exp\left(\simi{\mathbf{z}_i}{\mathbf{z}_p} / \tau \right)}{ \sum_{a \in I \setminus \{i\}} \exp\left(\simi{\mathbf{z}_i}{\mathbf{z}_a} / \tau\right)}
\end{align}
where 
$\tau \in \mathbb{R}^{+}$ is a scalar temperature parameter, 
$\simi{\mathbf{u}}{\mathbf{v}} = \frac{\mathbf{u} \cdot \mathbf{v}}{\|\mathbf{u}\| \, \|\mathbf{v}\|}$ is the cosine similarity metric %(\mathbf{u}/\|\mathbf{u}\|) \cdot (\mathbf{v}/\|\mathbf{v}\|)
and 
$P(i)$ is the set \emph{positive} examples with respect to the current \emph{anchor} $i$, %\in I := \{1, \dots, 2B\}
%i.e., all $|P(i)|$ samples in the same batch that belong to the same category as the anchor, out of the $C+R$ above-described categories.%3C
{i.e., all embeddings in the same batch that belong to the same category as the anchor, among the $C + R$ categories described above. We denote by $|P(i)|$ the number of such positive examples, i.e., the cardinality of $P(i)$. This loss is minimized by increasing the similarity between positive pairs (in the numerator) while decreasing the similarity between negative pairs (in the denominator).}

\subsection{Architecture details} \label{ssec:arch_details}

\paragraph{Positional encoder} The final positional embedding vector $\mathbf{p}$ is generated by a simple MLP
%The MLP that generates the final positional embedding vector $\mathbf{p}$, which we denoted $f_\theta$, is
composed of an input layer, a single hidden layer, and an output layer, with widths of $128$, $256$, and $128$, respectively. The fully connected layers, except the output one, are followed by batch normalization, ReLU (rectified linear unit) activation, and a dropout operation (active only during training). Finally, a sigmoid non-linearity is applied at the output layer. 
The input and output dimensions of this MLP being set to $128$, it implies that $2d=128$.

\paragraph{Branch encoders} The encoders of both region-invariant and region-specific branches consist of a 4-layer MLP with an input layer of size $(F+2d)$ followed by three layers of width 256. Here, every layer is followed by batch normalization, ReLU, and dropout operations. The output embeddings $\mathbf{z}_{inv}, \mathbf{z}_{spe}$ are, therefore, of size $D=256$. The input feature dimension is $F=109$ in our dataset, and the dropout rate is set to 50\%.

\paragraph{Classifiers} Both classifiers consist of a single linear layer with input dimension $D=256$ and output dimension given by the number of classes: $C$ for the land-cover classifier and 8 for the region classifier. The outputs are the logits, which, upon application of a softmax operation, serve as the input of the cross-entropy loss functions.

\section{Results}
\label{sec:results}
This section presents the results and analysis of our experimental evaluation conducted on the data set described in Section~\ref{sec:data}.

Our objective is to evaluate the performance of our framework, \method{}, across several dimensions. Firstly, a quantitative assessment is conducted to compare the performance of \method{} with baselines and competing approaches at both Level-1 (Table~\ref{tab:level1-classes}) and Level-2 (Table~\ref{tab:level2-classes}) LULC classification tasks. To this end, we examine two distinct scenarios: i) an Extrapolation scenario (\textit{EXTRAP}), where training data are available from all the biogeographical regions and, ii) a Leave-One-bioRegion-Out scenario (\textit{LORO}),
where we train the classification approaches using data from all biogeograpichal regions except one, and then they are tested on the leave-out region. The latter scenario is specifically designed to assess the spatial generalization capabilities of the competing approaches.
Secondly, we provide an in depth ablation analysis of \method{} in the LORO scenario. Then, we inspect the internal representations learned by our model, visually assessing the benefit to explicitly introduce geospatial information in the learning process. Lastly, we provide and discuss visual LULC map extracts generated by our framework.\footnote{The code associated with this paper is available at: \url{https://github.com/cassiofragadantas/LUCAS_LULC/}.}

\subsection{Competitors and experimental setting}
With the goal to assess the performance of \method{} in a large-scale land cover mapping setting, we selected three machine learning approaches commonly used in the field of remote sensing: Random Forest (RF), Extreme Gradient Boosting Machine (XGBoost), and Support Vector Machine (SVM). The choice of these approaches is also motivated by the nature of the data information described in Section~\ref{sec:data}.

In the \textit{EXTRAP} scenario, the $139,217$ available data samples are split into two parts: training and test sets following a proportion of $75\%$ and $25\%$ of the original data set, respectively. 

For the \textit{LORO} scenario, the splitting procedure is based on biogeographical regions. Given $L$ biogeographical regions, we iteratively select one region as the test set while using the remaining $L-1$ ones for training. This process is repeated $L$ times, with each region serving once as the test set. For this scenario, we report the average results across all iterations.

For each scenario, we consider two different levels of analysis. The first one, referred to as Level-1, involves a classification task among seven high-level LC classes, and the second one, referred to as Level-2, involves a crop-type classification task where only the agricultural crop classes are considered. This latter task involves $19$ for different crop-type classes. These two classification tasks enable us to assess the approaches under varying levels of thematic complexity. The Level-1 LC classification task represents a relatively easy scenario with distinct classes. In contrast, the Level-2 crop type classification task presents a more challenging setting, where the crop type classes exhibit high inter-class similarities, making them inherently more challenging to distinguish.

The assessment of the model performances was done considering the following metrics: \textit{Weighted F1-score} (simply indicated with \textit{F1-score}) and \textit{Accuracy} (global precision) score.

\subsubsection{Implementation details}

Experiments were carried out on a workstation with a dual Intel (R) Xeon (R) CPU E5-2667v4 (@3.20GHz) with 256 GB of RAM and TITAN X (Pascal) GPU. The proposed approach was implemented using the \textit{Pytorch} Deep Learning library \citep{Pytorch} and it runs on a single GPU. RF and SVM were implemented using the Python \textit{Scikit-learn} library~\citep{scikit-learn} and run on CPU. XGBoost was implemented using the XGBoost Python Package\footnote{\url{https://xgboost.readthedocs.io/en/latest/python}} and also ran on CPU.

Given the absence of a validation data set, we did not perform hyperparameter optimization for any of the compared methods, neither ours nor the competitors.
For our approach, the training stage was conducted for 500 epochs with a learning rate of 10$^{-4}$,  AdamW~\citep{LoshchilovH19} optimizer, and a batch size of 256.
For the RF classifiers, we employed the same settings as in \citet{ghassemi2024european}, that is, the number of trees in the forest set to 130, a minimum of 2 samples required at each leaf node, and all remaining parameters set to default Scikit-learn configuration with Gini impurity as the splitting criterion, no maximum depth limit for the trees and square root of the total number of features considered at each split. 
XGBoost and SVM classifiers were executed with the default configuration defined in the employed packages (XGBoost Python Package and Scikit-learn, respectively).
For XGBoost, it implied using a tree-based booster with a maximum depth of 6 and a uniform sampling method.
For SVM, it implied using radial basis function kernel with an L2 regularization parameter set to 1 and the kernel coefficient ($\gamma$) set to \emph{`scale'}, which adapts based on the data variance and number of features, i.e., \( \gamma = \frac{1}{\operatorname{Var}(\mathbf{X}) \cdot F} \).

\subsection{Extrapolation Scenario}
In this section, we report and analyze the results of the different competing methods under the \textit{EXTRAP} scenario. Table~\ref{tab:level1} shows the results, in terms of F1-Score and Accuracy, for the Level-1 (LC classification) and Level-2 (crop type classification) tasks, respectively. We can observe that \method{} consistently outperforms all the other approaches across both tasks. For the Level-1 task, all the approaches achieve F1-Scores within a narrow range of $77.68$ to $80.30$ percent, with less than three percent separating the best and worst performers. Regarding the Level-2 task, it reveals more substantial performance differences, with F1-Score spanning a wider range of around $8$ percent (from $55.83$ to $64.01$). Here, notably, \method{} overcomes the second-best performer (\textit{XGBoost}) by approximately $2.5$ percent of F1-Score. The Accuracy score exhibits similar performance patterns among the methods.

\begin{table}[!ht]
\centering
\caption{~Land cover classification (Level 1) and crop type classification (Level 2) average performances in terms of \textit{F1-Score} and \textit{Accuracy} (in $\%$) under the Extrapolation scenario. The best results and second-best results are highlighted in bold and underscored, respectively. \label{tab:level1}}
\begin{tabular}{l|c|c||c|c}
& \multicolumn{2}{c||}{Level-1} & \multicolumn{2}{c}{Level-2} \\ \hline

                      & \textbf{F1-score} & \textbf{Accuracy} & \textbf{F1-score} & \textbf{Accuracy}    \\ \hline
SVM                   & 77.68 & 78.03  & 55.83 &  58.25      \\
RF                    & 78.03  & 78.56 & 58.01 & 59.85      \\
XGBoost               & 79.33  & 79.75 & 61.58 & 62.88          \\
Ours                  & \textbf{80.30} & \textbf{80.49} & \textbf{64.01} & \textbf{64.24} \\ \hline
\end{tabular}
\end{table}

\begin{table*}[!ht]
\caption{~Per-class F1 scores (in $\%$) for Level 1 classification \label{tab:level1-perclass}. The best result per class is highlighted in bold, and the second best is underlined. Reported proportions correspond to the test set statistics.}
\centering
\begin{tabular}{l|ccccccc|c}
Class ID   & 0              & 1              & 2              & 3              & 4              & 5              & 6              & Weighted \\
Proportion (\%) & 35.1\%         & 27.0\%         & 0.7\%          & 0.6\%          & 10.6\%         & 0.2\%          & 25.8\%         & Average \\ \hline
SVM        & 83.86          & 72.06          & 19.75          & 27.23          & 57.13          & 30.21          & 86.59          & 77.68 \\
RF         & 84.17          & 72.12          & \underline{21.31} & \underline{50.32} & 57.45          & 30.56          & 86.79          & 78.03 \\
XGBoost    & \underline{84.95} & \underline{73.30} & 16.03          & \textbf{52.85} & \underline{60.88} & \underline{32.08} & \underline{88.16} & 79.33 \\

Ours       & \textbf{85.90} & \textbf{74.11} & \textbf{22.11} & 47.98          & \textbf{63.97} & \textbf{32.94} & \textbf{88.47} & \textbf{80.30} \\ 
\hline
\end{tabular}
\end{table*}

\begin{table*}[!ht]
\caption{~Per-class F1 scores (in $\%$) for Level 2 classification. Reported proportions correspond to the test set statistics. \label{tab:level2-perclass}}
\centering 
%\small %\small > \footnotesize > \scriptsize > \tiny
\normalsize
\begin{tabular}{l|ccccccccccccccccccc|c}
Class ID  &$ 0       $&$ 1       $&$ 2       $&$ 3       $&$ 4       $&$ 5       $&$ 6       $&$ 7 $&$ 8 $ &$ 9 $\\
Proportion ($\%$) & $ 8.3           $&$ 23.3          $&$ 2.6           $&$ 11.2          $&$ 2.9           $&$ 3.2           $&$ 15.2          $&$ 2.0  $ &$ 2.1$   &$ 0.7     $ \\ 
\hline
SVM        &$ 39.5          $&$ 62.9          $&$ 35.7          $&$ 39.3          $&$ 27.0          $&$ 28.1          $&$ 81.2          $&$ 66.1 $&$ 81.1 $  &$ 6.1           $\\
RF         &$ 42.3          $&$ 66.4          $&$ 39.5          $&$ 46.5          $&$ 29.4          $&$ 31.9          $&$ 78.3          $&$ 62.7          $&$ 76.5  $  &$ 6.1           $\\
XGBoost   &$ \textbf{46.2} $&$ 67.6    $&$ 37.8          $&$ 50.6          $&$ 31.0          $&$ 33.9          $&$ 85.2          $&$ \textbf{67.6}$&$ 82.7$  &$ 21.1          $\\
Ours       &$ \underline{44.8}    $&$ \textbf{70.8} $&$ \textbf{47.6} $&$ \textbf{52.4} $&$ \textbf{35.2} $&$ \textbf{37.5} $&$ \textbf{86.7} $&$ \underline{66.7}          $&$ \textbf{84.1} $ &$ \textbf{31.1}    $\\
\hline
\hline
Class ID  &$ 10 $&$ 11 $&$ 12 $&$ 13 $&$ 14 $&$ 15 $&$ 16 $&$ 17 $&$ 18$& Wtd.\\
Proportion ($\%$) &$ 7.3      $&$ 2.0      $&$ 3.1      $&$ 7.5      $&$ 0.6      $&$ 4.0      $&$ 1.2      $&$ 2.7      $&$ 0.1 $ & Avg. \\
\hline
SVM     &$ 74.3          $&$ 19.1          $&$ 34.2          $&$ 66.5          $&$ 20.0          $&$ 67.9          $&$ 50.4          $&$ 6.5           $&$ 33.3   $ & $55.83$\\
RF      &$ 73.6          $&$ 33.2          $&$ 33.8          $&$ 66.9          $&$ 18.9          $&$ 66.8          $&$ 52.2          $&$ 22.0          $&$ 33.3$ & $58.01$\\
XGBoost    &$ 74.6          $&$ 37.8          $&$ 39.5          $&$ 69.6          $&$ 11.6          $&$ 74.8          $&$ \textbf{61.6}    $&$ 27.3          $&$ 33.3 $& $61.58$\\
Ours       &$ \textbf{75.3} $&$ \textbf{44.8} $&$ \textbf{42.2}    $&$ \textbf{72.3} $&$ \textbf{25.7}    $&$ \textbf{78.1}    $&$ \underline{61.2} $&$ \textbf{30.8}    $&$ \textbf{54.6}$& $ \textbf{64.01}$\\
\hline
\end{tabular}
\end{table*}

\begin{figure}[h!]
    \centering
    \begin{subfigure}[b]{0.49\textwidth}    
        \includegraphics[width=\linewidth]{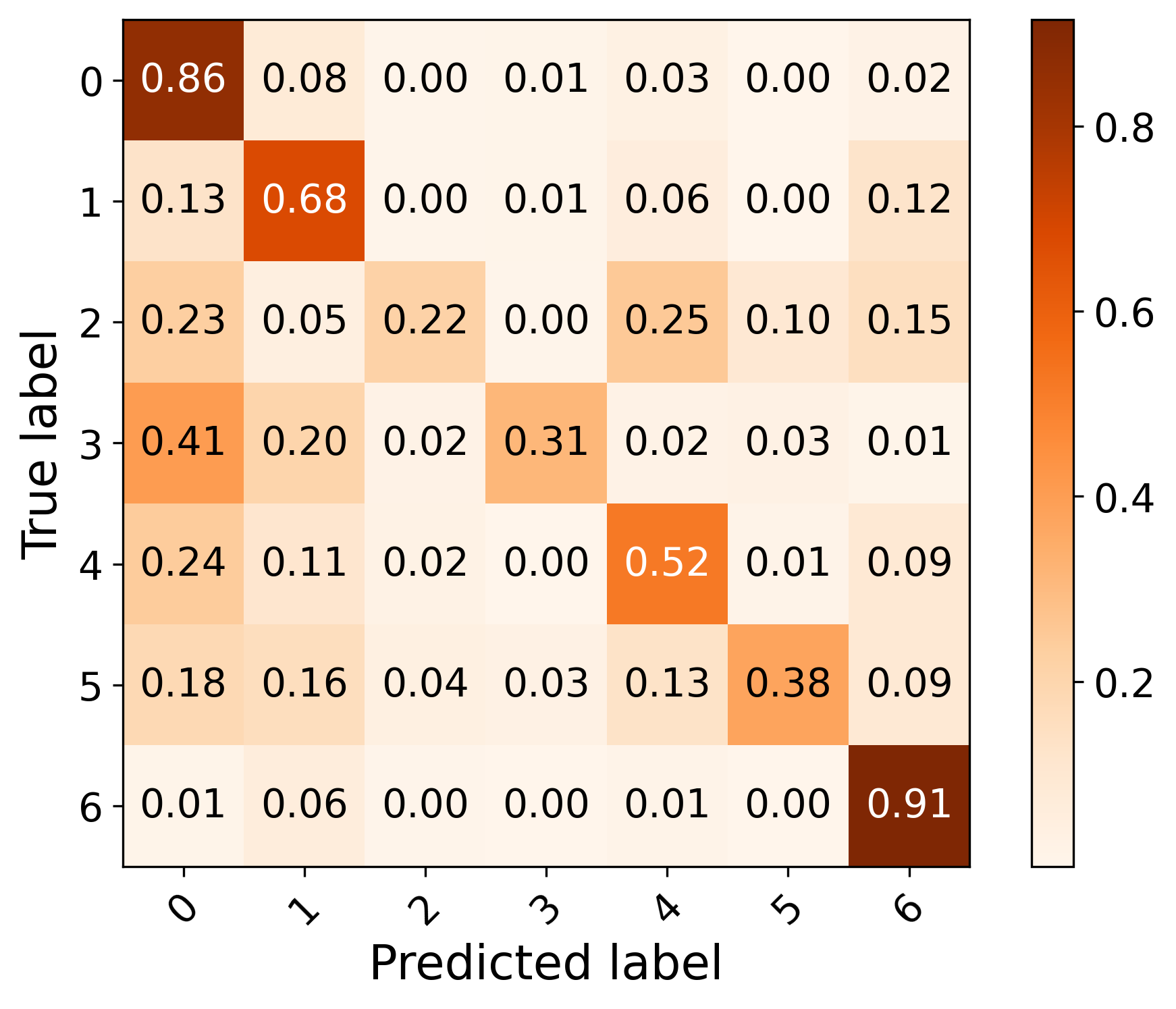}
        \caption{SVM}
    \end{subfigure}
    \hfill    
    \begin{subfigure}[b]{0.49\textwidth}
        \centering
        \includegraphics[width=\linewidth]{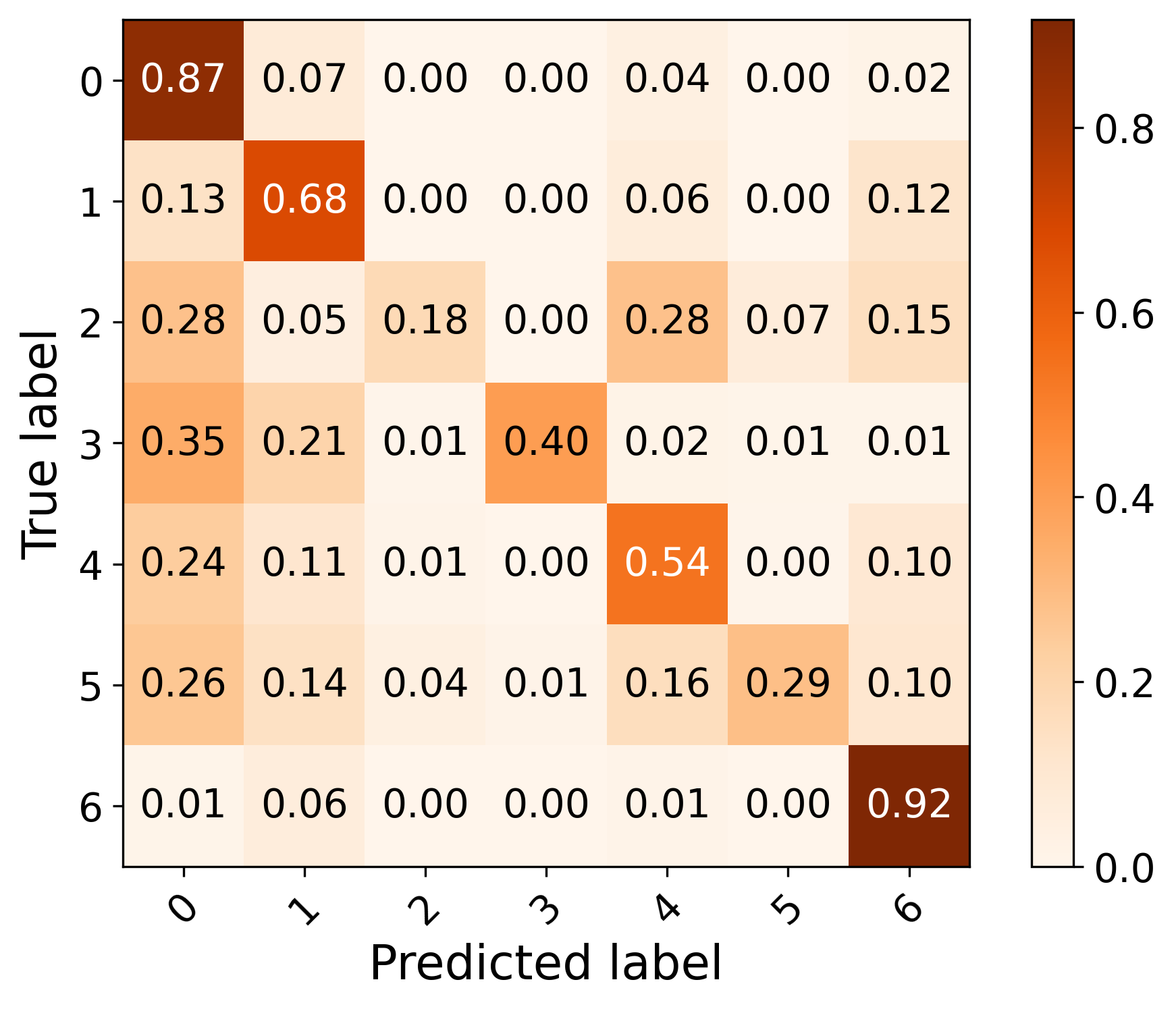}
        \caption{RF}
    \end{subfigure}
    \vskip\baselineskip
    \begin{subfigure}[b]{0.49\textwidth}
        \centering
        \includegraphics[width=\linewidth]{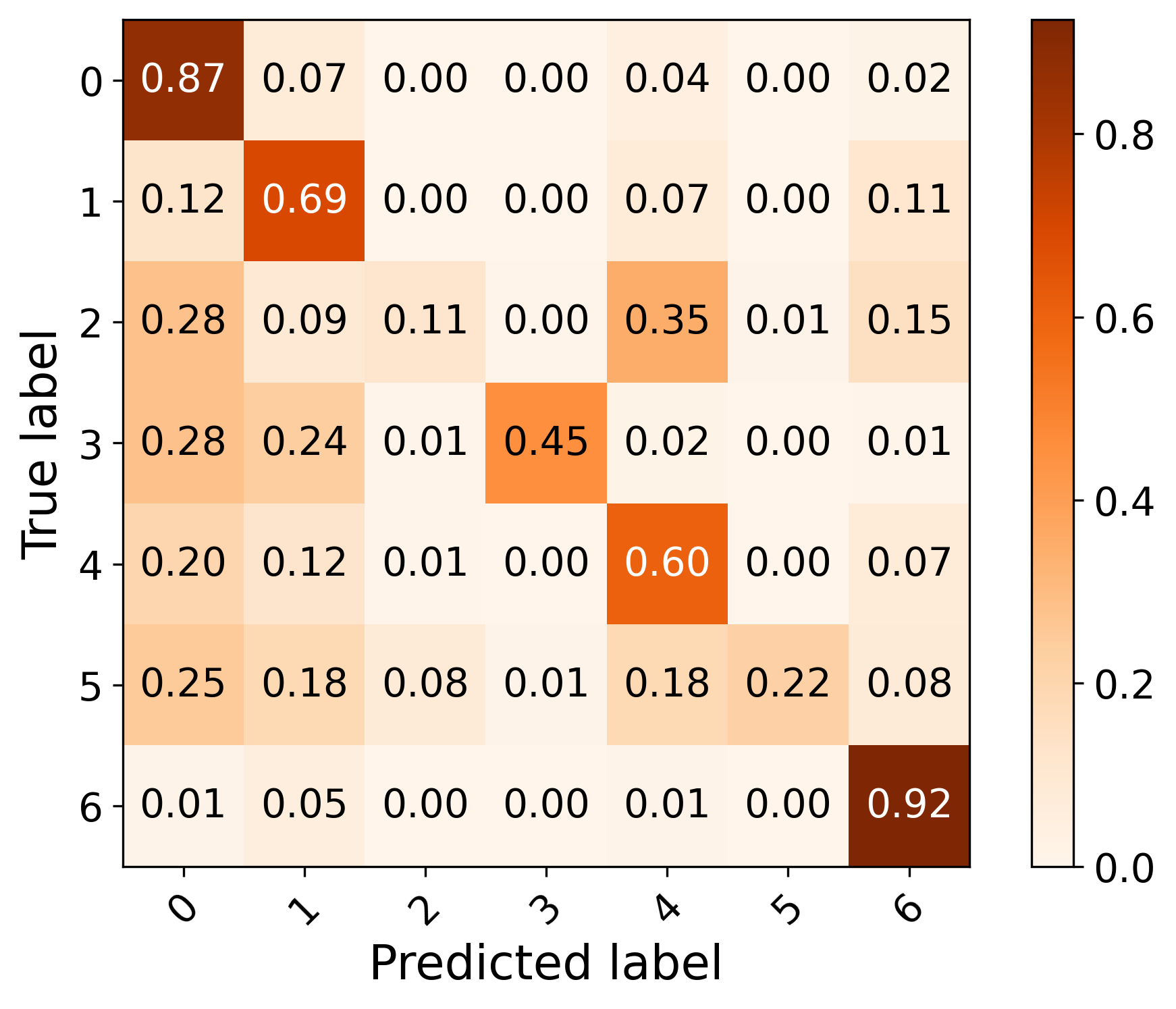}
        \caption{XGBoost}
    \end{subfigure}
    \hfill    
    \begin{subfigure}[b]{0.49\textwidth}    
        \includegraphics[width=\linewidth]{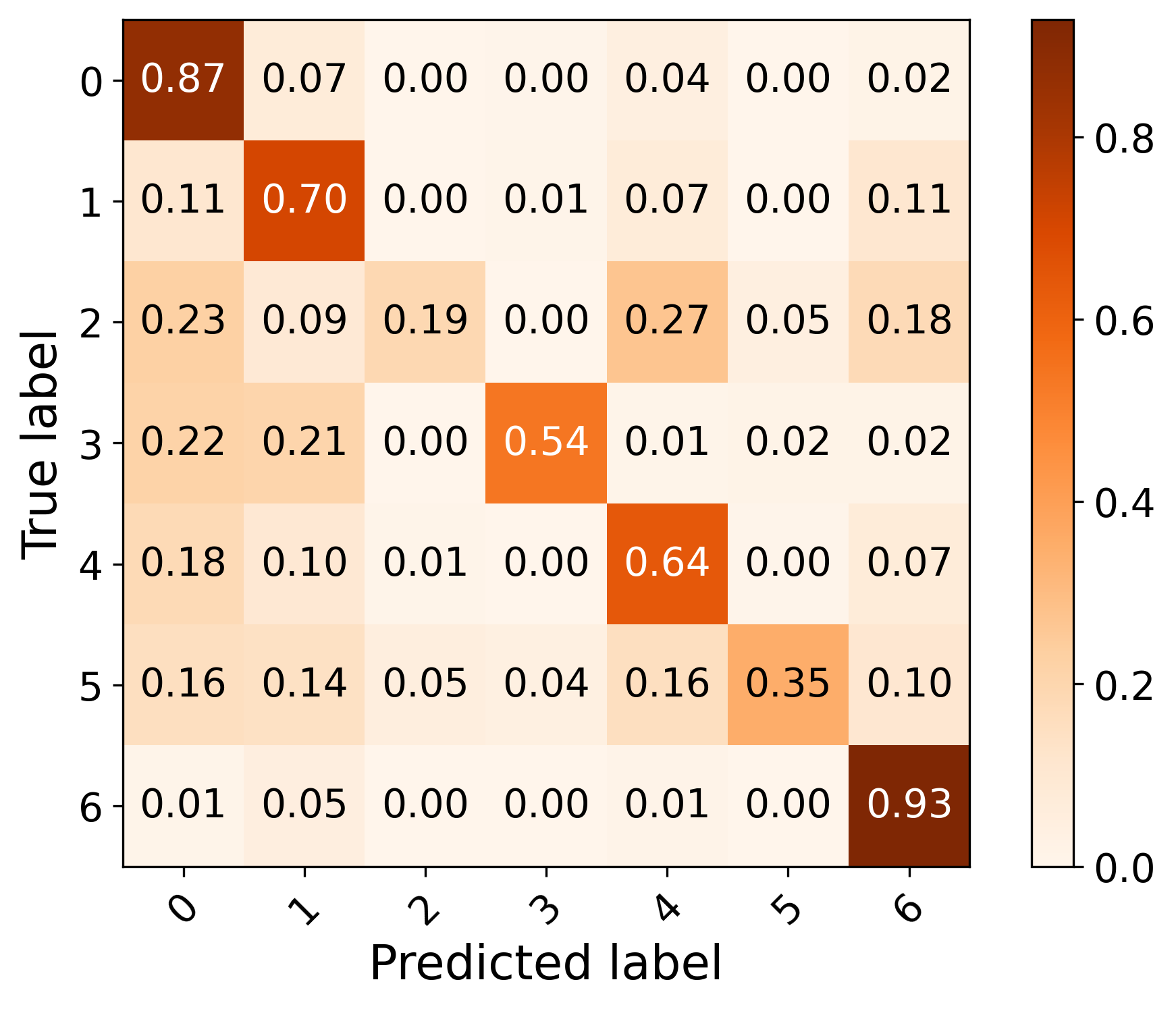}
        \caption{Ours}
    \end{subfigure}
    \caption{Confusion matrices (row-normalized) for Level-1 classification, extrapolation scenario.}
    \label{fig:Confusion-lev1}
\end{figure}

\begin{figure}[h!]
    \centering
    \begin{subfigure}[b]{0.49\textwidth}    
        \includegraphics[width=\linewidth]{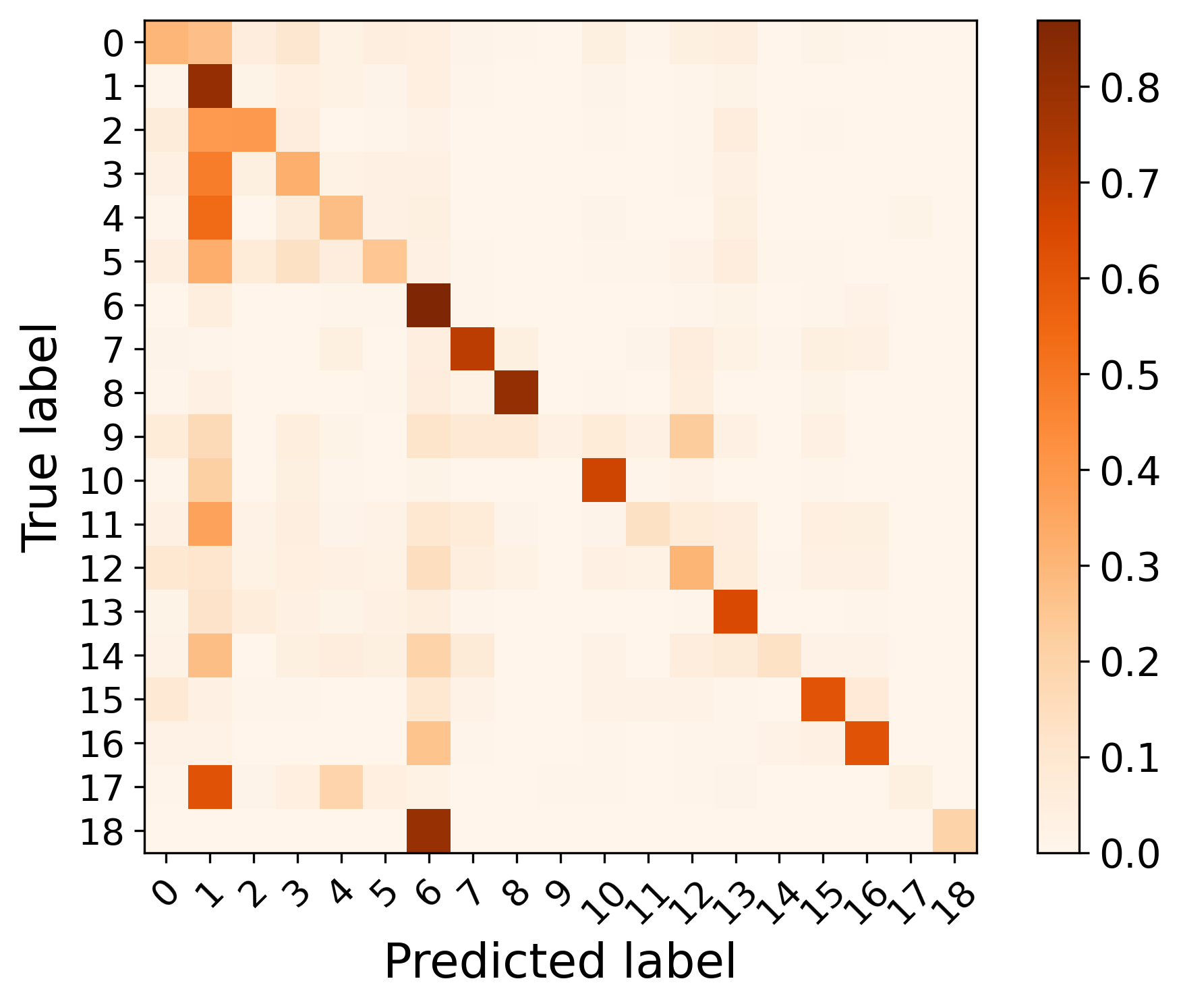}
        \caption{SVM}
    \end{subfigure}
    \hfill    
    \begin{subfigure}[b]{0.49\textwidth}
        \centering
        \includegraphics[width=\linewidth]{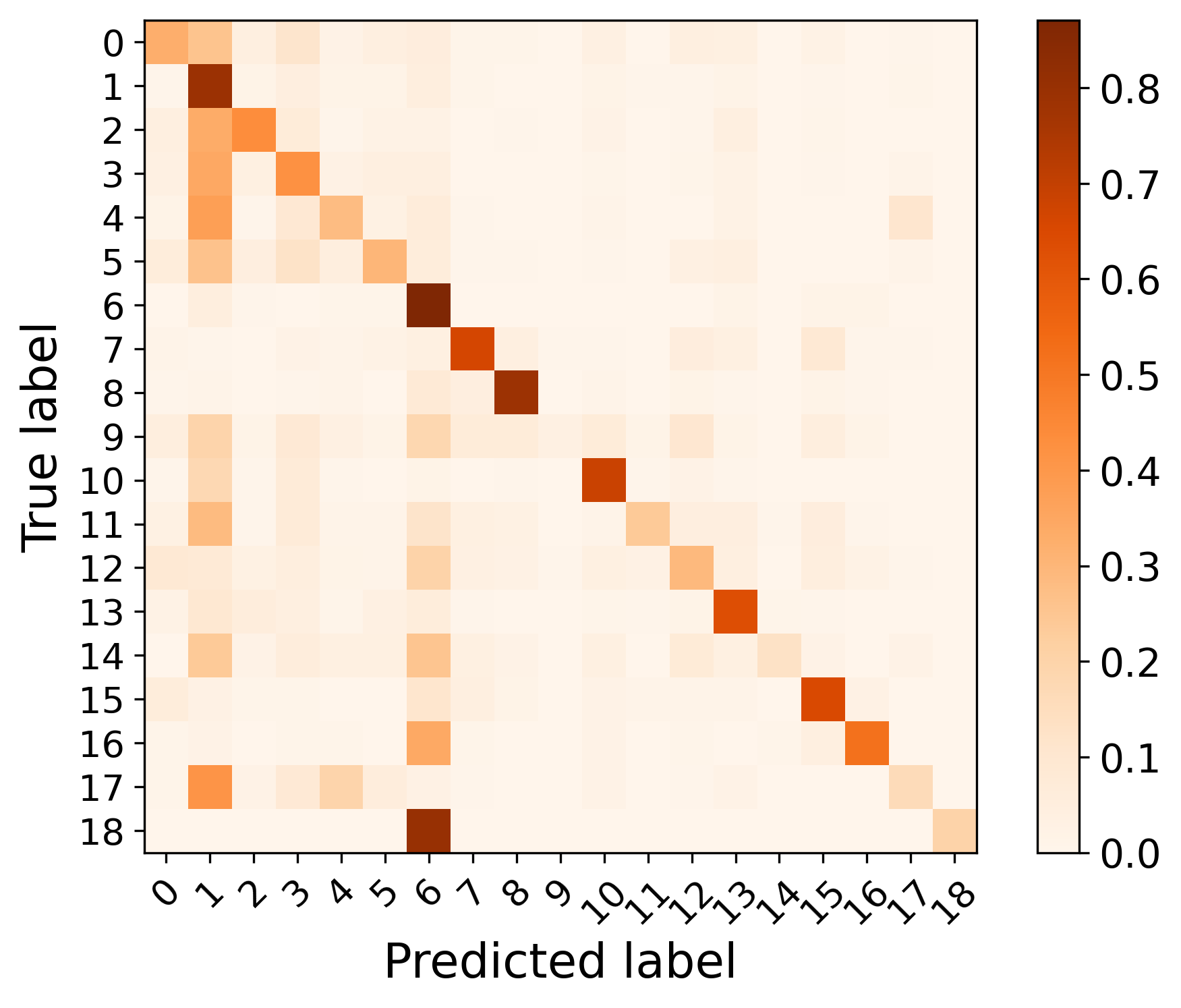}
        \caption{RF}
    \end{subfigure}
    \vskip\baselineskip
    \begin{subfigure}[b]{0.49\textwidth}
        \centering
        \includegraphics[width=\linewidth]{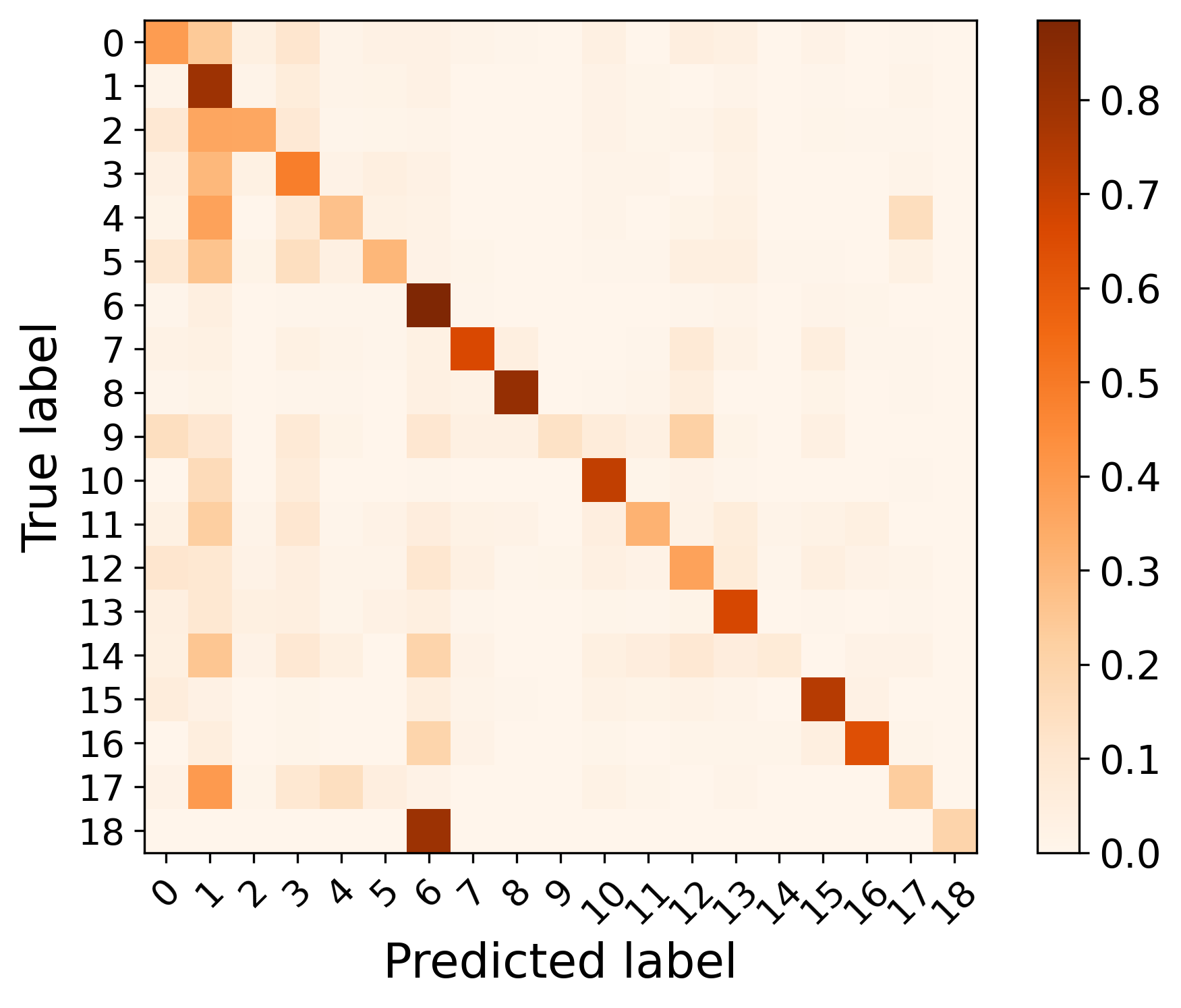}
        \caption{XGBoost}
    \end{subfigure}
    \hfill    
    \begin{subfigure}[b]{0.49\textwidth}    
        \includegraphics[width=\linewidth]{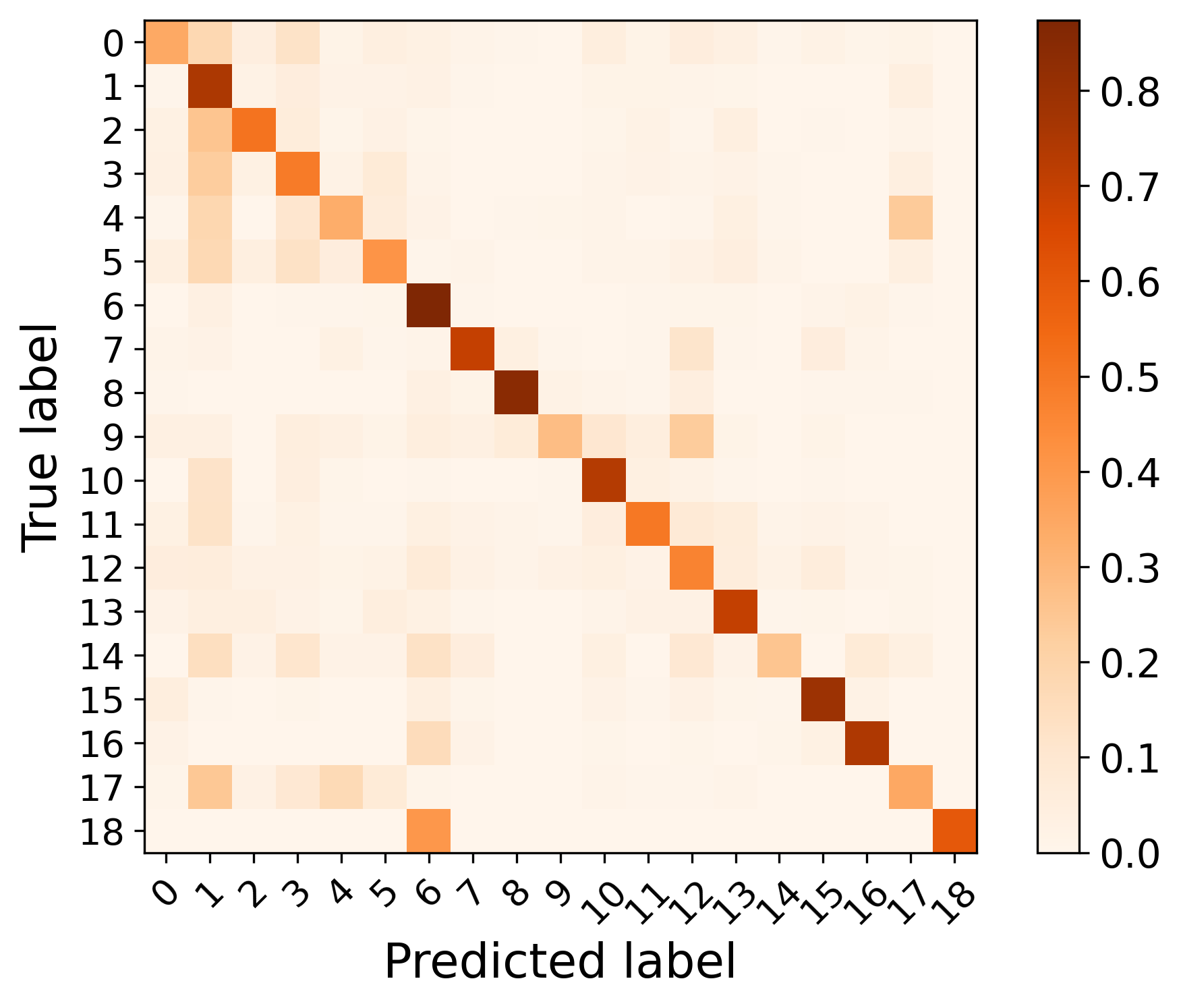}
        \caption{Ours}
    \end{subfigure}
    \caption{Confusion matrices (row-normalized) for Level-2 classification, extrapolation scenario.}
    \label{fig:Confusion-lev2}
\end{figure}

Table~\ref{tab:level1-perclass} and Table~\ref{tab:level2-perclass} report the per class F1-Score performances for the Level-1 and Level-2 classification tasks, respectively. Regarding the LC classification task (Table~\ref{tab:level1-perclass}), our proposed approach achieves either the best or second best performance across all classes, with the exception of class 3 (Wetlands). Generally, \method{} shows less pronounced advantages in underrepresented classes (those comprising less than $1\%$ of the data set), where limited data availability constrains the learning process. Concerning the crop type mapping task (Table~\ref{tab:level2-perclass}), \method{} outperforms all the competing approaches on $16$ classes over $19$ while exhibiting the best-second performances on the remaining three crop type classes (Bare arable land, Potatoes, and Soya).

The results obtained from the Extrapolation scenario clearly demonstrate that incorporating multi-level geospatial context (both geographical coordinates and biogeographical region information) enhances the performance of land cover classification and crop type mapping at a continental scale. This improvement is particularly evidenced in the crop type classification, as underlined by the wide performance variation among competing approaches and the pronounced performance gap between \method{} and competitors that do not use geospatial information. These findings reinforce the fundamental relationship between agricultural crop phenology and local environmental and climatic conditions. The results underscore that integrating such geospatial information is crucial for improving the effectiveness of large-scale LULC mapping frameworks.

\subsubsection {Confusion matrices} 

Class confusions for the extrapolation scenario test set are summarized in Figures \ref{fig:Confusion-lev1} and \ref{fig:Confusion-lev2}, respectively, for Level-1 and Level-2 classification. The reported confusion matrices are normalized by row.
In Figure \ref{fig:Confusion-lev1} we can see that class 2 (Bare land and lichens/moss) is a challenging class as its samples are commonly misclassified, being most often misidentified as classes 4 (Artificial land) or 0 (Woodland). Two other challenging classes are Wetlands (class 3) and Water (class 5); however, notice that our proposed approach is capable of improving upon the discriminability of such classes, especially reducing the confusion between Wetlands (class 3) and Woodland (class 0) significantly. By inspecting the columns of the matrices, we can see that class 0 (Woodland) and, to a lesser degree, class 1 (Grassland) are a common source of false positive predictions (i.e., samples from other classes are likely misclassified as one of these two classes). Yet, notice how this phenomenon is less pronounced in the proposed approach. When it comes to crop classification (Level-2) in Figure \ref{fig:Confusion-lev2}, we can see that the off-diagonal elements are significantly less pronounced in the confusion matrix associated to our proposed approach. Particularly, we highlight the reader's attention to the confusion associated with columns of classes 1 and 6 (i.e., samples from other classes that are wrongly classified as one of these classes), which are noticeably reduced by our method.

\subsection{Leave-One-bioRegion-Out scenario}

In this section, we report and analyze the performance of the competing methods under the \textit{LORO} scenario. Table~\ref{tab:LORO_summary} presents the results, in terms of F1-Score and Accuracy, for both Level-1 (LC classification) and Level-2 (crop type classification) tasks, respectively. As expected, performance metrics in the \textit{LORO} scenario are lower than in \textit{EXTRAP} scenario, as models need to generalize on samples from a biogeographical region that was not present during the training stage. In this challenging setting, \method{} consistently outperforms all competing approaches across both tasks, achieving an average improvement of nearly two points across all metrics. Similar to the \textit{EXTRAP} scenario, the performance gap between methods is more pronounced in crop type classification than in LC classification. Table~\ref{tab:LORO_lev1} and Table~\ref{tab:LORO_lev2} report detailed F1-Score performances, per region, for Level-1 and Level-2 tasks, respectively. For LC classification (Table~\ref{tab:LORO_lev1}), \method{} achieves the highest performance in six out of eight regions. In the remaining two regions (\textit{Black Sea} and \textit{Steppic}), the limited number of test samples may affect the reliable evaluation of different methods. Nevertheless, \method{} maintains competitive performance compared to other approaches in these regions.
For crop-type classification (Table~\ref{tab:LORO_lev2}), \method{} demonstrates superior performance across most spatial generalization cases. In the two exceptions (\textit{Black Sea} and \textit{Continental}), our approach still achieves the second-best performance among all methods.
The Leave-One-bioRegion-Out results demonstrate that \method{}'s integration of multi-level geospatial information still provides benefits even when calibration data from certain regions is unavailable. These findings highlight once more the importance of incorporating geospatial information for improving large-scale LULC mapping frameworks.

\begin{table}[!ht]
\centering
\caption{~Land cover classification (Level 1) and crop type classification (Level 2) average performances in terms of {F1-Score} and {Accuracy} (in $\%$) under the Leave-one-region-out (LORO) scenario. The displayed results are average results over all eight regions. The best results and second-best results are highlighted in bold and underscored, respectively. \label{tab:LORO_summary}}
\begin{tabular}{l|c|c||c|c}
& \multicolumn{2}{c||}{Level-1} & \multicolumn{2}{c}{Level-2} \\ \hline

                      & \textbf{F1-score} & \textbf{Accuracy} & \textbf{F1-score} & \textbf{Accuracy}    \\ \hline
SVM                   & $71.94$ & $72.93$  & $47.45$ &  $52.18$      \\
RF                    & $70.10$  & $72.15$ & $48.14$ & $52.93$      \\
XGBoost               & $71.15$  & $72.98$ & $52.88$ & $56.70$          \\
Ours                  & $\textbf{73.59}$ & $\textbf{74.77}$ &$ \textbf{54.44}$ & $\textbf{58.29}$ \\ \hline
\end{tabular}
\end{table}

\begin{table*}[!ht]
\caption{~Weighted F1 scores for Leave-One-bioRegion-Out (LORO) level 1 classification. \label{tab:LORO_lev1}}
\centering
\begin{tabular}{l|cccc}
 Region                & Alpine & Atlantic & Black Sea & Boreal \\ 
Percentage (\%) & 4.8\%	& 15.5\% &	0.2\% &	13.7\%  \\
                            \hline
SVM                         & \underline{72.59}  & 78.7     & \underline{86.66} & 73.75    \\
RF                          & 71.57  & 77.76    & \textbf{88.54}     & 72.16  \\
XGBoost                     & 70.03  & \underline{79.35}    & 88.4      & \underline{74.36}   \\
Ours & \textbf{76.82}  &  \textbf{79.72}    & 86.4      &  \textbf{74.96}  \\
\hline 
\hline
Region             & Continental & Mediterranean & Pannonian & Steppic \\
Percentage ($\%$)&	39.2\% &	22.6\% &	3.4\% &	0.6\%\\
   \hline
SVM                         & \underline{77.91}       & \underline{55.16}  & \underline{61.96} & \textbf{68.78} \\
RF                          & 77.39       & 53.18         & 59.41     & 60.75   \\
XGBoost                     & 77.78       & 54.45         & 60.16     & 64.63  \\
Ours &  \textbf{79.06}       &  \textbf{57.89}         &  \textbf{65.86}     & \underline{67.98} \\
\hline
\end{tabular}
\end{table*}

\begin{table*}[]
\caption{~Weighted F1 scores ($\%$) for Leave-One-bioRegion-Out (LORO) level 2 classification. Column names stand for the held-out region. \label{tab:LORO_lev2}}
\centering
\begin{tabular}{l|cccccccc}
Region                            & Alpine & Atlantic & Black Sea & Boreal  \\ 
Percentage (\%) & 1.8\%	 &	20.7\%	& 0.3\%		& 5.7\%		 \\ \hline
SVM                         & 47.92  & 50.03    & 74.63     & 33.48   \\
RF                          & 46.85  & 50.15    & 74.5      & 39.79   \\
XGBoost     & \underline{53.53}  & \underline{54.92}    & \textbf{78.67}     & \underline{40.97}     \\
Ours &  \textbf{62.28}  &  \textbf{57.29}    & \underline{75.61}     & \textbf{42.17}   \\
\hline
\hline
Region & Continental & Medit. & Pannonian & Steppic \\
Percentage (\%) & 50.9\% & 16.1\% & 3.6\%	& 0.9\%	\\
\hline
SVM & 40.17       & 15.61         & 52.97     & 64.77 \\
RF & 40.82       & 16.11         & 52.81     & 64.11  \\
XGBoost & \textbf{45.66}       & \underline{25.09}         & \underline{55.96} & \underline{68.2} \\
Ours &  \underline{44.91}       &  \textbf{26.15}         &  \textbf{57.92}     &  \textbf{69.18}  \\
\hline
\end{tabular}
\end{table*}

\subsection{Ablation analysis}

Table~\ref{tab:ablation} presents an ablation analysis of the proposed framework, examining the impact of both fine-grained (latitude/longitude) and coarse-grained (biogeographical region) geospatial information under the Leave-One-bioRegion-Out (LORO) scenario.

We compare two ways of integrating the latitude/longitude information: (i) using fixed (non-learnable) positional encodings (PEs) in equations \eqref{eq:pos_enc1}--\eqref{eq:pos_enc2} and (ii) using our learnable PE approach where an additional MLP block, learned in an end-to-end fashion, further processes the coordinate-based fixed encodings. Results show a significant performance drop when using fixed positional encodings alone, especially compared to the learned PE variant. This degradation indicates that directly incorporating fine-grained geolocation information might not be helpful for the downstream classification task and, in fact, can even hinder its performance. By contrast, our learnable PE mechanism, allows the additional MLP model to learn how to properly represent and utilize spatial information in a way that seems more appropriate 
for the downstream classification task.

Our analysis reveals that incorporating either latitude/longitude (with learned positional encodings) or biogeographical region information enhances classification performance across all tasks compared to the baseline model, which lacks any geolocation information (first row in Table~\ref{tab:ablation}).
One can note that biogeographical region information provides benefits for Level-1 (LC classification) tasks, whereas latitude/longitude information proves to be more influential for Level-2 (crop type classification) tasks.
Finally, the joint exploitation of both fine-grained and coarse-grained geospatial information, as in our proposed framework (last row), yields, on average, improved classification results across nearly all situations.

%% MULTICLASS DISENTANGLEMENT - Leave-one-region-out average perf

\begin{table}[!ht]
\caption{~Ablation analysis of the proposed approach to assessing the contribution of fine grain (lat-long) and coarse grain (biogeographical region) geospatial information during the learning process. The results are obtained under the Leave-One-bioRegion-Out (LORO) scenario. \label{tab:ablation}}
\centering
\begin{tabular}{ccc||cc||cc}
Lat-long & Learnt & Biogeographical region & \multicolumn{2}{c||}{Level 1}     & \multicolumn{2}{c}{Level 2}     \\
                             information & PE & information & \textbf{Acc.}  & \textbf{F1}    & \textbf{Acc.}  & \textbf{F1}    \\ \hline
 & & & 74.49          & 73.17         & 57.76          & 53.93          \\
 & & \checkmark & 74.59          & 73.26        & 58.18          & 54.37          \\
 \checkmark &  &  &  71.90	 & 70.20 & 55.70 & 	52.14  \\
 \checkmark &  & \checkmark & 70.93	& 69.18  &  55.61 & 52.21 \\ 
 \checkmark & \checkmark & & 74.47          & 73.03          & 58.27          & \textbf{54.88}          \\
 \checkmark & \checkmark & \checkmark & \textbf{74.77} & \textbf{73.59} & \textbf{58.29} & 54.44 \\ 
\end{tabular}
\end{table}

\subsection{Examples of land cover mapping}
\begin{figure*}[p]
    \centering
    \includegraphics[width=\textwidth]{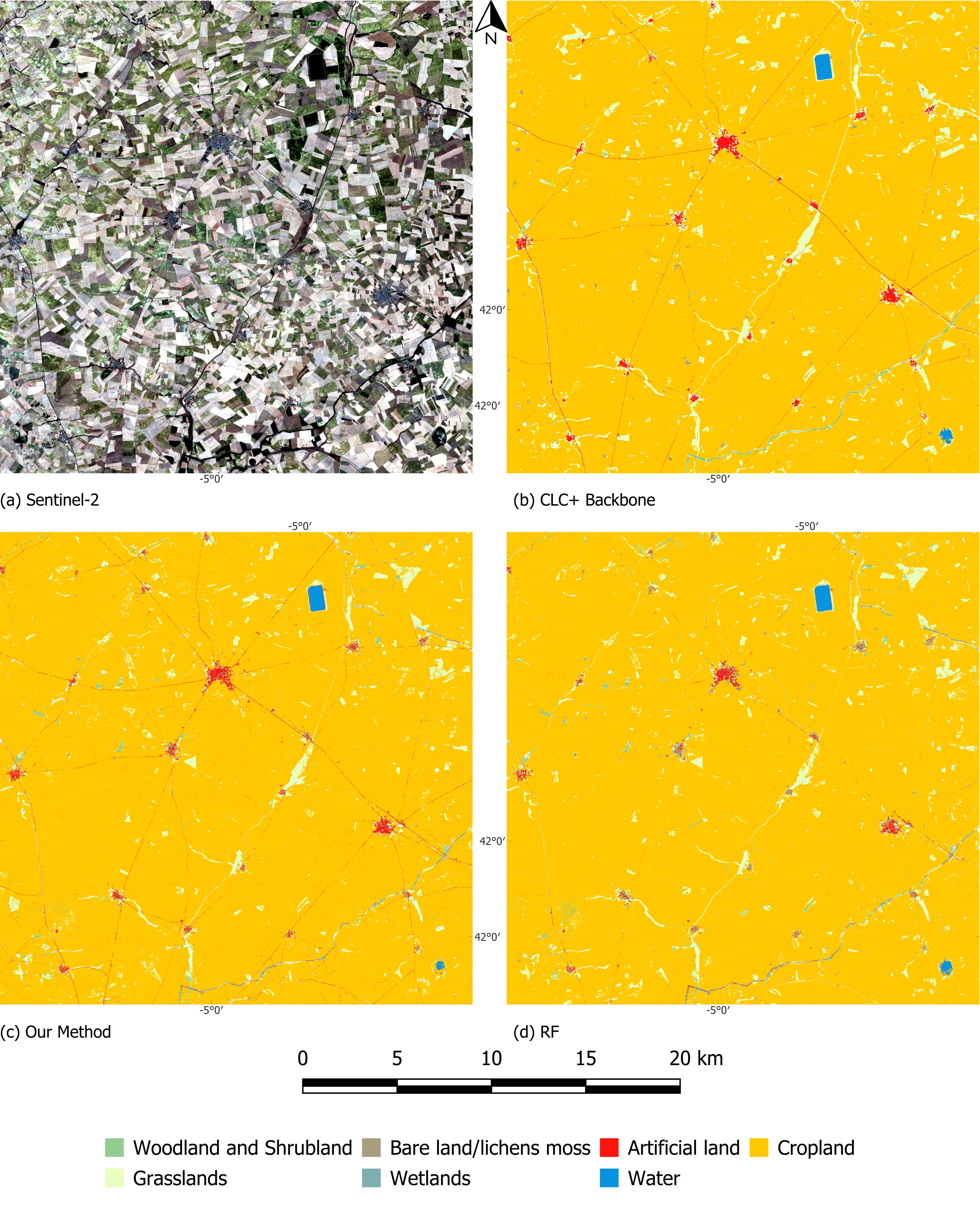}
    \caption{LC maps Spain area - Mediterranean region.}
    \label{fig:LCmap-Spain}
\end{figure*}

\begin{figure*}[p]
    \centering
    \includegraphics[width=\textwidth]{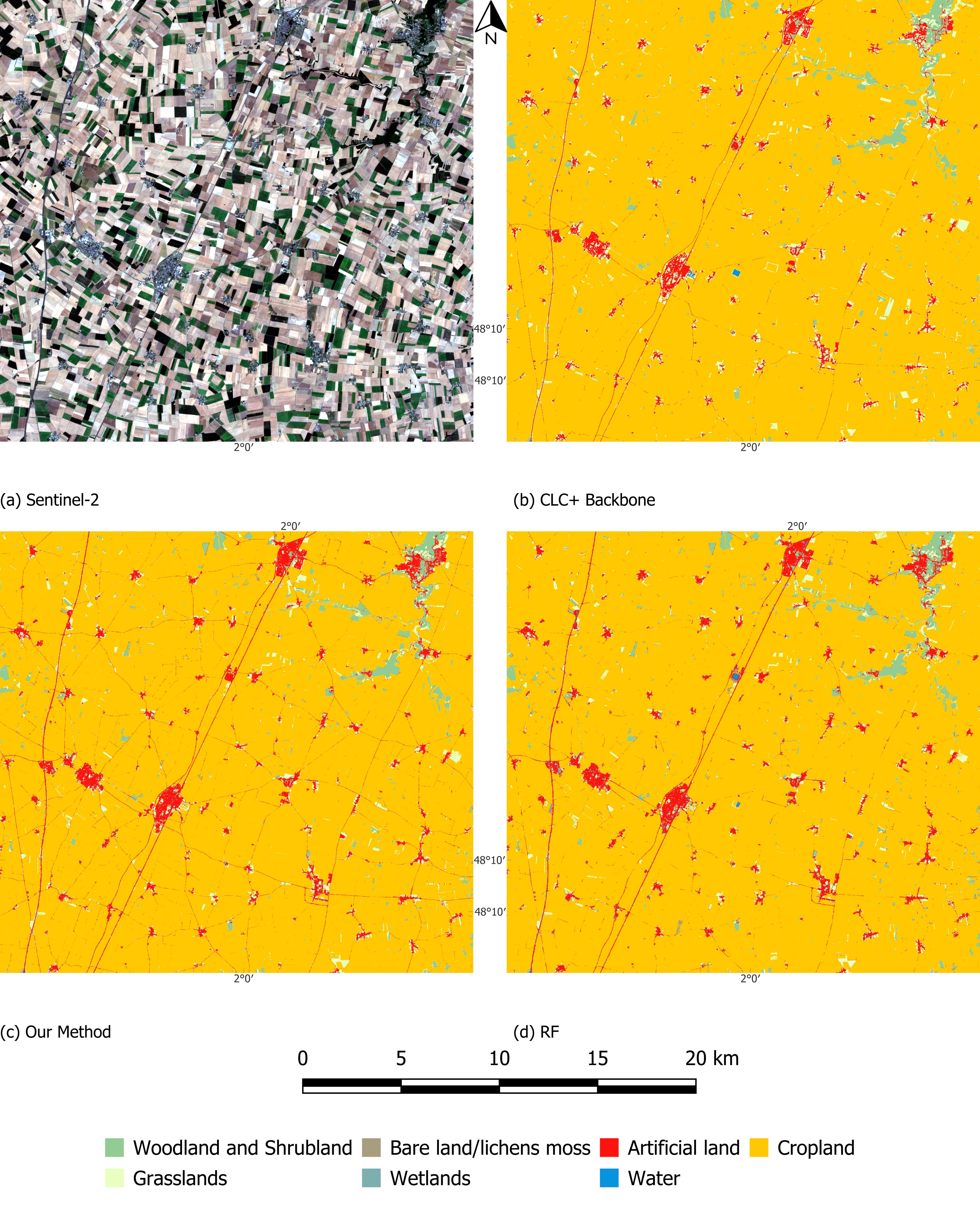}
    \caption{LC maps France area - Atlantic region.}
    \label{fig:LCmap-France}
\end{figure*}

\begin{figure*}[p]
    \centering
    \includegraphics[width=\textwidth]{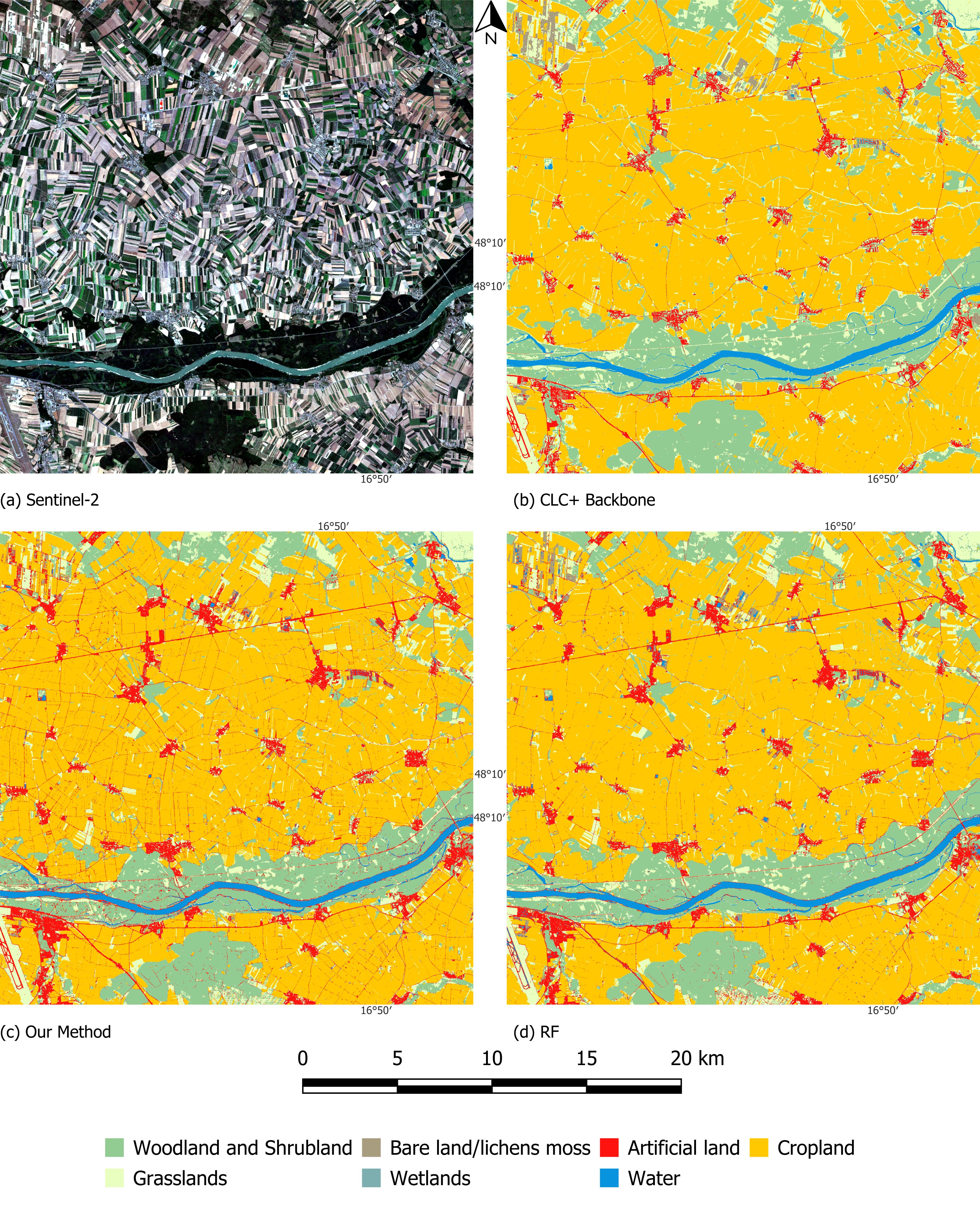}
    \caption{LC maps Austria area - Continental region.}
    \label{fig:LCmap-Austria}
\end{figure*}

\begin{figure*}[p]
    \centering
    \includegraphics[width=\textwidth]{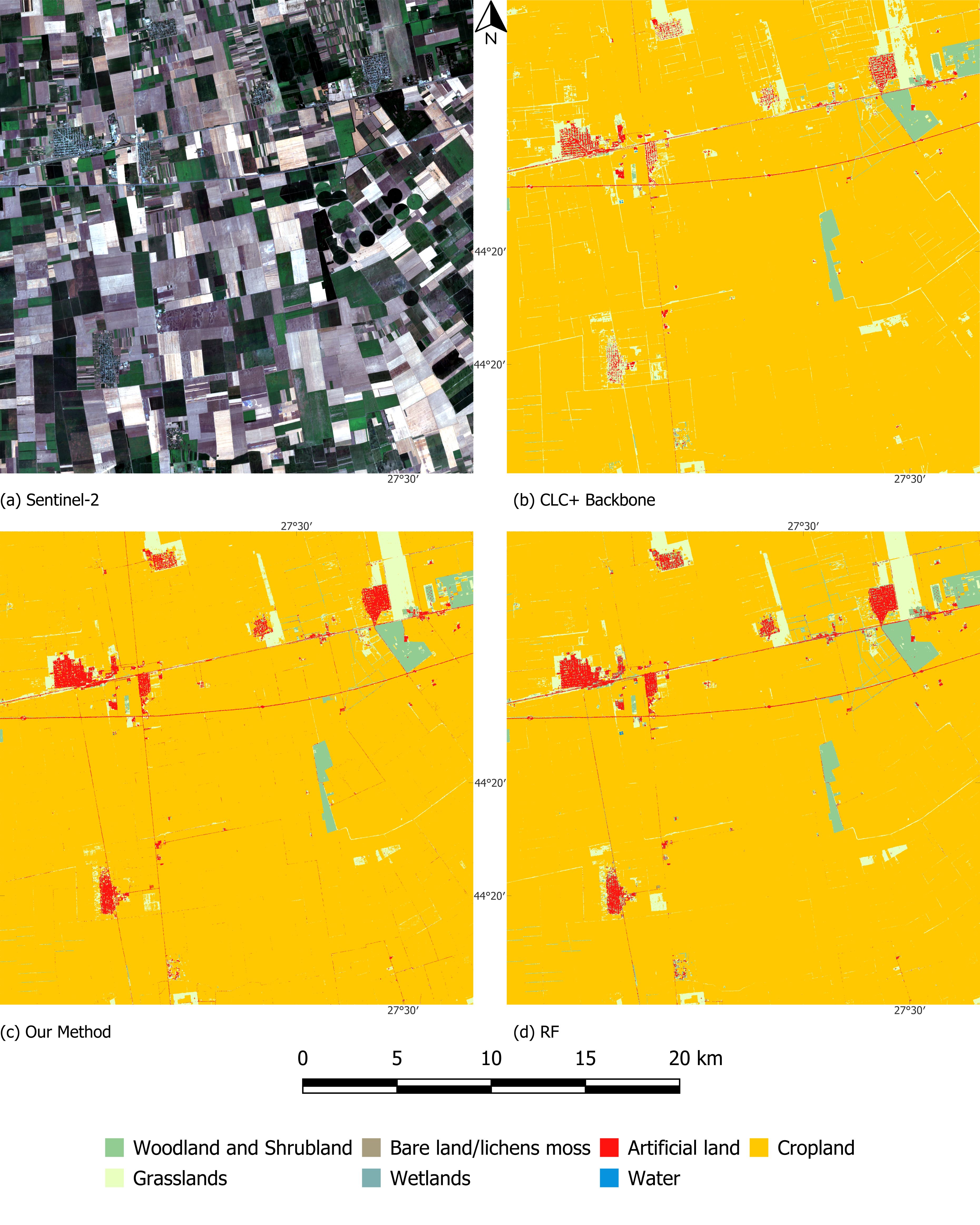}
    \caption{LC maps Romania area - Steppic region.}
    \label{fig:LCmap-Romania}
\end{figure*}

In this section, we provide land cover maps for the following four sites located in different biogeographical regions, with an area of 25 $\text{km}^2$, each of them: 
\begin{enumerate}
    \item Spain - Castile and León (Mediterranean) [Figure~\ref{fig:LCmap-Spain}]
    \item France - Centre-Val de Loire (Atlantic) [Figure~\ref{fig:LCmap-France}]
    \item Austria - Lower Austria (Continental) [Figure~\ref{fig:LCmap-Austria}]
    \item Romania - Sud-Muntenia (Steppic) [Figure~\ref{fig:LCmap-Romania}]
\end{enumerate}

For each site, four different maps are provided: (a) S2 satellite imagery, (b) the Corine Land Cover plus Backbone (CLC+ Backbone)~\footnote{\url{https://land.copernicus.eu/en/products/clc-backbone}} map, (c) the land cover (LC) map produced by our method, and (d) the map generated using RF.

As no accurate Europe-wide LULC map for 2022 is currently available, we use the CLC+ Backbone data set from 2021 \footnote{\url{https://doi.org/10.2909/71fc9d1b-479f-4da1-aa66-662a2fff2cf7 }} as a reference for comparison. Produced by the EEA, this dataset offers 10 m spatial resolution, covers Europe with 11 dominant LC classes, and is updated every three years with an overall accuracy of $93.7\%$. The CLC+ class labels were aggregated as following steps to align with our classes to ensure a direct comparison:

\begin{itemize}  
    \item All tree-related classes (Woody needle-leaved trees, Woody Broadleaved deciduous trees, Woody Broadleaved evergreen trees, and  Low-growing woody plants) were merged into Woodland/Shrubland.  
    \item Permanent herbaceous and Periodically herbaceous classes were relabeled as Grassland and Cropland, respectively. 
    \item Sealed surfaces were relabeled as Artificial land, while Water retained its original label.  
\end{itemize}  

The LC maps generated by our method exhibit greater detail and fragmentation, which reflects a more precise and realistic representation of the landscape across various biogeographical zones. In contrast to existing methods, which simplify land cover distributions, our approach effectively captures small-scale variations and heterogeneous patterns. Detail improvement is particularly noticeable in transition areas, such as forest and grassland edges, and at agricultural boundaries, where subtle differences in land cover are better reflected. Additionally, our method significantly improves the extraction of linear features, such as roads and narrow land cover conversions, which are often misclassified by traditional methods. This enhancement is especially important in fragmented landscapes, where the clear delineation of roads, rivers, and field boundaries is essential for accurate land cover mapping. Our model provides an accurate and granular representation of terrain by effectively distinguishing the abovementioned features, which makes it an excellent choice for applications involving detailed spatial analysis.

\subsection{Inspection of the proposed approach}

To gain a deeper insight into the behavior of the proposed model, we conducted a visual inspection of its internal representations and learned spatial embeddings. This analysis aims to complement the quantitative evaluations presented earlier by providing intuitive, qualitative evidence of how incorporating spatial information—both coarse and fine-grained—affects the structure of learned features. Specifically, we explore two aspects: 
\begin{enumerate}
    \item the distribution of internal model's features via t-SNE projection (\Cref{ssec:tsne});
    \item the structure of the learned location embeddings through PCA-based RGB visualization (\Cref{ssec:locEmb}).
\end{enumerate}

\begin{figure}[p]
    \centering
    \begin{subfigure}[b]{0.9\textwidth}
        \centering
        \includegraphics[width=0.49\linewidth]{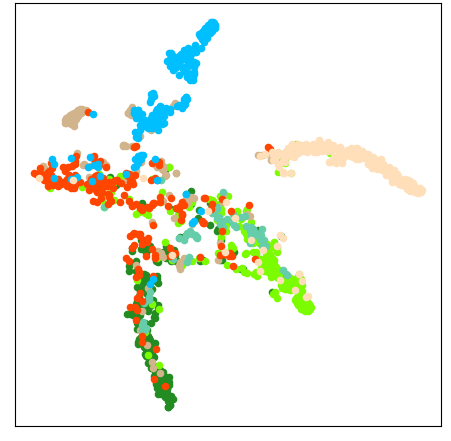}
        \hfill
        \includegraphics[width=0.49\linewidth]{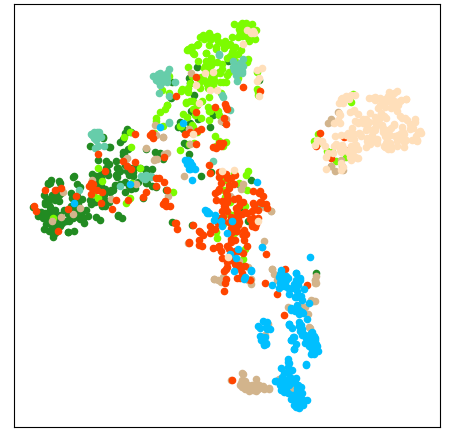}
        \caption{Alpine region on test (LORO)}
    \end{subfigure}
    \vskip\baselineskip
    \begin{subfigure}[b]{0.9\textwidth}
        \centering
        \includegraphics[width=0.49\linewidth]{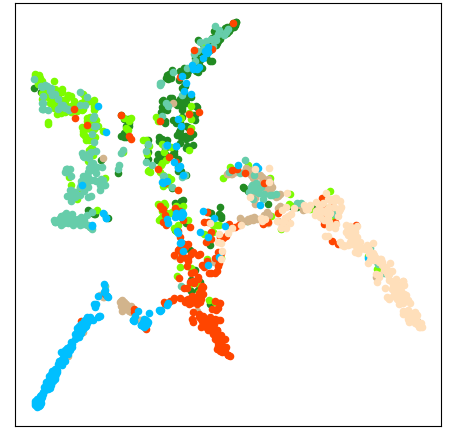}
        \hfill
        \centering
        \includegraphics[width=0.49\linewidth]{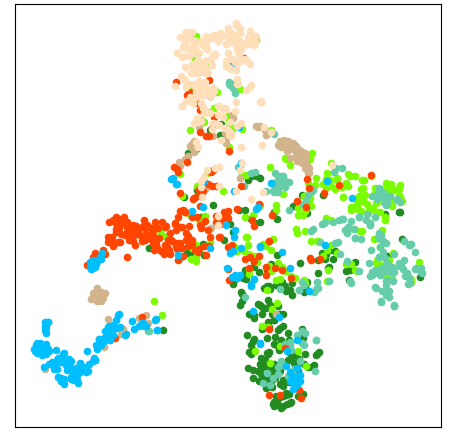}
        \caption{Pannonian region on test (LORO)}
    \end{subfigure}    
    % Single image below the grid
    \vskip\baselineskip
    \includegraphics[width=0.7\textwidth]{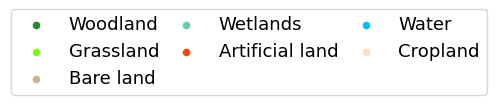}
    \caption{t-SNE visualization of the embeddings extracted by the proposed approach (left column) and its ablated version without coarse and fine-grained spatial information (right column) considering the Leave-One-bioRegion-Out scenario on the Alpine (top row) and the Pannonian (bottom row) regions.}
    \label{fig:tsne}
\end{figure}

\subsubsection{Visualisation of internal model representations}
\label{ssec:tsne}
In Figure~\ref{fig:tsne}, we provide a visual inspection of the internal feature representation learned by \method{} compared to an ablated version of the proposed model that excludes both coarse and fine-grained spatial information. Our analysis focuses on the Leave-One-bioRegion-Out (LORO) scenario, where we visualize the internal feature representations of biogeographical regions that were not used during training. To this end, we randomly selected 50 samples from each land cover class and extracted their corresponding feature representations for each method. We then applied t-SNE~\citep{Maaten2008VisualizingDU} to reduce the feature dimensionality for visualisation purposes.
As shown in Figure~\ref{fig:tsne}, the feature representations learned by \method{} demonstrate a more compact and well-defined curvilinear cluster structure compared to the ablated version, showing reduced visual confusion between some of the land cover classes. For instance, in the Alpine region (top row), the $Water$ and $Bare$ $land$ classes are clearly separated in our method but remain mixed in the ablated version. These visualization results align with our previously discussed quantitative and qualitative findings.

\begin{figure}[p]
    \centering
    \begin{subfigure}[b]{0.49\textwidth}
        \centering
        \includegraphics[width=\linewidth]{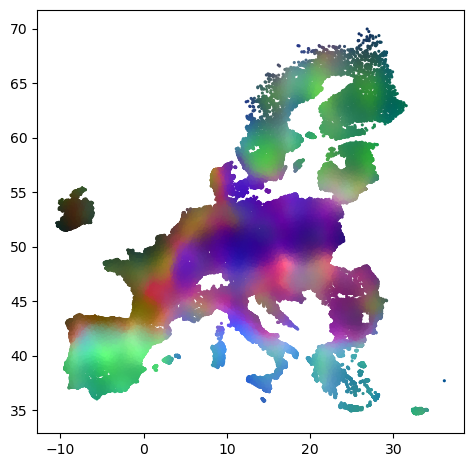}
        \caption{Level 1 - coarse + fine-grained info}
    \end{subfigure}
    \hfill
    \begin{subfigure}[b]{0.49\textwidth}
        \centering
        \includegraphics[width=\linewidth]{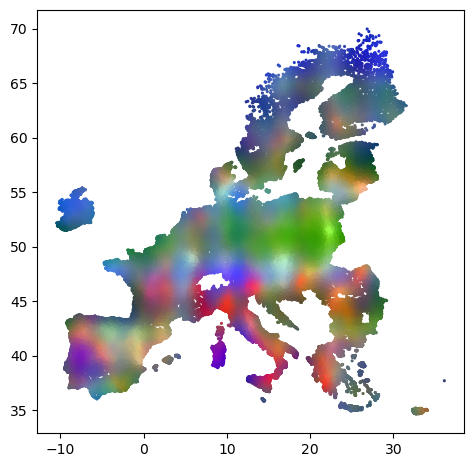}
        \caption{Level 1 - fine-grained info only}
    \end{subfigure}
    \vskip\baselineskip
    \begin{subfigure}[b]{0.49\textwidth}
        \centering
        \includegraphics[width=\linewidth]{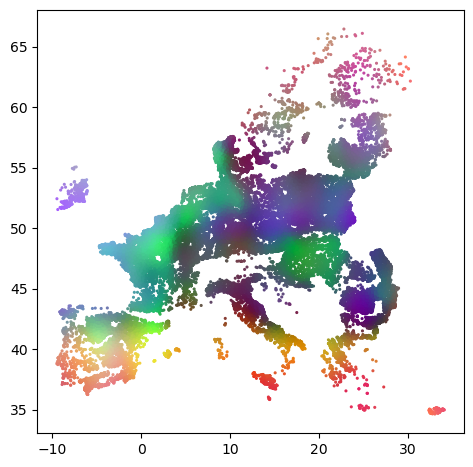}
        \caption{Level 2 - coarse + fine-grained info}
    \end{subfigure}
    \hfill
    \begin{subfigure}[b]{0.49\textwidth}
        \centering
        \includegraphics[width=\linewidth]{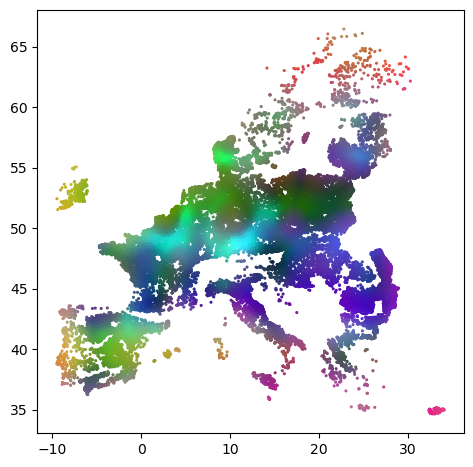}
        \caption{Level 2 - fine-grained info only}
    \end{subfigure}
    \caption{Visualization of the learned location embeddings when both coarse and fine-grained spatial information are exploited (left column) and when only the fine-grained information is used (right column).} 
    \label{fig:locEmb}
\end{figure}

\subsubsection{Visualisation of the learned location embeddings}
\label{ssec:locEmb}
The use of both coarse and fine-grained spatial information during the training of \method{} enables the deep neural network to learn geographically informed data representations. To investigate how our framework integrates such spatial knowledge, following recent approaches focused on extracting generic spatially-aware feature representations~\citep{russwurm2023geographic,mai2023sphere2vec,bourcier2025learning}, we visualize the positional representation learnt by \method{}. To this end, we extracted the positional representation of each sample of our dataset and applied PCA dimensionality reduction to obtain a three-dimensional representation. We then interpreted these three dimensions as RGB values to assign distinct colors to each sample.

Figure~\ref{fig:locEmb} shows the positional embedding visualization for Level 1 (top row) and Level 2 (bottom row), comparing \method{} (left column) with a version of our framework that only considers fine-grained information (right column). At both classification levels, we can observe that the use of biogeographical region information yields a smoother representation with extensive areas exhibiting homogeneous color. 
Conversely, the ablated version, which only considers fine-grained information, exhibits clear visual discontinuities with abrupt transitions between small, patchy regions. This pattern suggests an over-specialization behaviour of the ablation to highly localized geographical areas. Furthermore, we can also note that \method{} implicitly recovers the biogeographical region structure used in the disentanglement process, while this is not the case for the ablated model. This phenomenon is particularly evident in the Level 2 (crop type) classification task (bottom line), where \method{} assigns similar positional representations (orange/red color) across the Mediterranean area (southern Spain, Italy, and Greece), while, the ablated model generates, for the same area, distinctly different positional representations, as indicated by the diverse, uncorrelated colors.

\section{Discussion}
\label{sec:discuss}
This study shows how the integration of base spatial information with DL techniques can create innovative ways to significantly improve large-scale LULC classification across the EU-27 using a broad (Level-1 with 7 LULC classes) and a more detailed (Level-2 with 19 crop type classes) classification scheme. It discusses the credibility of EO data, the efficiency of ground truth data, the overall classification pipeline, the analysis of results, and the advantages of the proposed method over traditional approaches. A robust framework for accurate LULC mapping was proposed by integrating fine-grained (latitude-longitude coordinates) and coarse-grained (biogeographical regions) spatial information combined with advanced DL techniques.

\subsection{Earth Observation and Ground Truth Data}

Combining S1 and S2 data with ancillary data sets creates a strong foundation for land cover classification. By merging optical and radar data, we can capture a wide range of land surface features while reducing problems caused by cloud cover in S2 data. With ancillary data, such as LST and DEM, the model is further able to capture environmental and topographical variations. It is worth mentioning that, following \citet{ghassemi2024european}, a subset of $109$ effective features out of available $196$ were selected for the final classification, in which only annual percentile information obtained from S2 data were used in the classification process in order to maintain data consistency across the EU-27. While this approach supports large-scale consistency, accuracy could potentially be improved by using temporally richer (e.g., monthly) data, albeit at the cost of reduced spatial coverage due to cloud-induced data gaps.

The ground truth data, derived from the LUCAS 2022 data set, plays a vital role in training and testing the classification models. This data set is enhanced significantly, and its efficiency is improved by the Copernicus module by adding polygons that represent uniform land cover areas. A total of $139,217$ samples were used after training data balancing procedures were carried out on $81,427$ primary samples.

Land cover samples that are labeled across EU-27 allow for comprehensive analysis, which reduces bias caused by region-specific data sets. Nevertheless, there are still challenges with underrepresented classes, especially those with low sample availability, which can affect generalization.
The final model shows the capability of handling both broad land cover and specific crop types classification, as LUCAS data is defined in two different Level-1 (7 classes) and Level-2 (19 classes) schemes. Besides performing a hierarchical classification, utilizing a mixture of those schemes is able to make the model versatile.

\subsection{Classification Methodology}

Our new classification framework integrates fine-grained (latitude-longitude coordinates) and coarse-grained (biogeographical regions) spatial information into a deep learning-based MLP architecture. Using positional encoding and biogeographical partitioning, the model can capture spatial patterns and regional variations in land cover. 

The training procedure utilizes a dual-branch architecture, with one branch dedicated to land cover classification and the other to region-specific classification. Using contrastive loss, the model effectively discriminates between different land cover classes while taking into account the geographical context. The use of supervised contrastive loss enriches the embedding space, which improves the model's ability to discriminate between samples of different classes.

The positional encoding module converts latitude and longitude coordinates into a fixed-length vector (128 dimensions) using a sinusoidal function inspired by Transformer models. This vector is then passed through an MLP to generate a learned positional encoder. This later is concatenated with spectral bands and indices as detailed in Section~\ref{sec:data}. While the region-specific branch generates embeddings that capture the biogeographic region of the sample, the land cover classification branch focuses on the land cover classes. The feature disentanglement approach allows the model to separate region-specific information from region-invariant information. Consequently, it helps to gain better generalization capabilities.

\subsection{Classification Results Analysis}

According to the classification results obtained under the two scenarios EXTRAP (training data are available from all the biogeographical regions) and LORO (training data from all regions except one are used and then tested on the leave-out region), the framework can efficiently perform LULC classification tasks, especially for crop type classification (Level-2).

The scope of the EXTRAP scenario is to evaluate the model’s ability to classify land cover types within known geographic distributions. The proposed method achieves an average F1-score of 80.30\% for Level-1 classification and 64.01\% for Level-2 classification, surpassing all baseline models. Besides, our method shows notable improvement compared to XGBoost,
which achieves an F1-score of 79.33\% and 61.58\% for Level-1 and Level-2, respectively. The accuracy scores follow a similar trend, in which we achieved 80.49\% and 64.24\% accuracy for Level-1 and Level-2, individually, whereas XGBoost achieved 79.75\% and 62.88\% accuracy. The confusion matrices indicate that the method effectively reduces misclassification errors, particularly for classes Durum wheat (F1-score: 47.6\%), Rye (F1-score: 35.2\%), Other roots crops (F1-score: 31.1\%), Other non-permanent industrial crops (F1-score: 44.8\%), Other cereals (F1-score: 25.7\%), and Rice (F1-score: 54.6\%).

The LORO scenario aims to evaluate how well the model generalises to unseen biogeographical regions, as each region is systematically omitted during training and used only for testing. The presented model outperforms all competitors at Level-1 and Level-2, achieving an average F1 score of 73.59\% and 54.44\%, respectively. Similarly, our method achieved 74.77\% accuracy for Level-1 and 58.29\% accuracy for Level-2. The performance gaps are larger for Level-2 classification, demonstrating the robustness of the model in identifying crop types in unseen landscapes.

The performance of classification varies by region. Due to its complex vegetation and agricultural patterns, the Mediterranean region presents significant challenges, resulting in lower F1-scores of 57.89\% for Level-1 and 26.15\% overall. The Continental region, in contrast, which covers 39.2\% of the study area, exhibits the highest F1 score at 79.06\% for Level-1 and 44.91\% for Level-2. The proposed model outperforms traditional methods, even though it faces challenges when applied to new biogeographical regions. Although the model achieves high F1-scores across most land cover classes, it struggles to classify underrepresented categories like Wetlands and Bare Land due to their similar spectral properties compared to other classes (i.e., Woodland and Artificial land, respectively).

Compared to RF and XGBoost, the developed method reduces misclassification errors more effectively, particularly for complex categories such as Artificial land (F1-score: 63.97\%) and Cropland (F1-score: 88.47\%). The developed framework produces more coherent and spatially consistent outputs, particularly in heterogeneous regions, based on spatial visualization.

The ablation analysis further demonstrates the contributions of geospatial metadata. Compared to the traditional models without geospatial inputs, latitude-longitude coordinates, and biogeographical region information enhances classification performance across all tasks. The results indicate that the biogeographical region improves Level-1 classification, increasing the F1-score from 73.17\% to 73.26\%. Meanwhile, latitude-longitude coordinates contribute more to Level-2 classification accuracy by improving the F1-score from 53.93\% to 54.88\%. It is noteworthy that coupling fine-grained and coarse-grained geospatial information yields the best overall performance, with accuracy and F1-scores of 74.77\%, 73.59\%, and 58.29\%, 54.44\%, respectively. 

The results demonstrate the model's capacity to enhance classification accuracy in both known and unseen geographical areas. These findings highlight the significance of integrating geospatial information into classification frameworks for large-scale mapping, which proves the model's suitability for extensive LULC mapping tasks.

\subsection{Advantages of the Proposed Method Over Traditional Approaches}

Our method outperforms traditional ML approaches such as RF, SVM, and XGBoost in both Level-1 and Level-2 classification. A key factor is the integration of geospatial metadata - latitude-longitude coordinates and biogeographic information - which improves accuracy and generalization, as demonstrated in the LORO scenario. The model mitigates overfitting by separating region-specific from region-invariant features. Despite using DL techniques, it remains computationally efficient and serves as a viable alternative to RF and XGBoost. Our results show consistently higher F1-scores and accuracy than baseline methods, confirming its robustness. The ablation analysis highlights the complementary role of fine- and coarse-grained spatial information: biogeographic data slightly improves Level-1 performance, while latitude-longitude coordinates significantly affect Level-2. Their combined use yields the highest accuracy, demonstrating their synergistic effect.

\subsection{The visual interpretation of the output maps}

The visual interpretation of the generated LULC maps across four distinct biogeographical regions demonstrates that our method offers a more detailed and realistic representation of land cover. Compared to traditional methods, this improvement is particularly evident in heterogeneous and transitional landscapes, as well as linear objects. According to maps provided by our method, the model effectively captures small-scale variations and preserves fine land cover structures that are often lost by conventional classification approaches. This is especially clear at the edges of forests and agricultural boundaries, where subtle changes in land cover are more evidently outlined. The improved extraction of linear features, such as roads and narrow land cover transitions, also highlights the model’s ability to distinguish small but critical landscape elements accurately. This visual analysis suggests our method can provide more dynamic and up-to-date LULC maps by offering a more granular and adaptive classification.

\section{Conclusions}
\label{sec:conclu}
 In the face of current global challenges, the importance of regular and accurate land cover classification cannot be overstated. For example, political instability often leads to changes in land use, which can have significant impacts on food security and the environment. Accurate land cover classification helps policymakers and environmentalists monitor these changes and assess their effects on ecosystems. Climate change exacerbates the need for precise land cover data. As global temperatures rise, natural disasters such as wildfires, floods, and droughts become more frequent and severe. Accurate land cover classification enables better prediction and management of these events by providing essential information on vegetation, water bodies, and urban areas. Moreover, land cover data is crucial for tracking deforestation, urban sprawl, and agricultural land use. These activities contribute to greenhouse gas emissions and biodiversity loss, further intensifying the climate crisis. By regularly updating land cover classifications, we can better understand and mitigate these impacts, ensuring more sustainable land use practices.

A critical assessment of raw imagery data, ground truth, and ancillary data such as digital elevation models (DEMs), climate data, and land use records, combined with domain-specific knowledge, is essential to improve the detail and accuracy of land cover classification. The availability of frequent and high-quality observations from the Copernicus Sentinel-1 and Sentinel-2 platforms, together with the Landsat mission, has significantly enhanced the ability to map land use and land cover (LULC) with increased temporal revisit time, spatial resolution, and thematic details. Additionally, the advent of cloud computing infrastructures has revolutionized data access and processing. Examples include Google Earth Engine, the Microsoft Planetary Computer, and the Sentinel Hub, as well as open-source platforms with unified application programming interfaces (APIs) for geospatial processing, such as OpenEO. Concurrently, efficient machine learning algorithms, such as Random Forest, have enabled both local and large-scale classification efforts, making mapping tools and results more accessible to a wider audience. Efforts in developing innovative classification algorithms, the availability of extensive ground truth data (e.g., LUCAS survey data for Europe), and a more operational perspective on continental (e.g., European CLCplus Backbone map) or global (e.g., ESA WorldCereal map) classification workflows and products exemplify these advances.

Recently, innovative deep learning (DL) methods have emerged, bringing improvements to the challenge of land cover classification. However, the literature suggests that classification algorithms alone are unlikely to drive a paradigm shift in achievable accuracy, especially for large-scale (e.g., continental or global) and thematically detailed land cover maps. 
Besides, while DL methods can achieve higher accuracy than traditional algorithms like RF, the improvements are often marginal and come at a high computational cost. 
For example, DL models like SITS-BERT can provide accuracy up to 91.3\% \(\sim \) 98.8\% in different regions, in comparison to 87.9\% \(\sim \) 94.21\% for RF models, but require more computational resources \citep{Yuan2020SelfSupervisedPO}. Similarly, the PAN method benefiting from TSNet models achieves an average of 85.7\% accuracy over three different sites, while RF models deliver an average of 82.5\% \citep{WANG2022Cross}. Another study shows CNN on EO data reaching 94.6\% as opposed to 88.7\% for RF \citep{Kussul2017Deep}. Utilizing the U-Net method on S2 data, an accuracy of 97.8\% was achieved, while the performance of RF was 96.2\% \citep{Singh2022Deep}.
Interdisciplinary skills and perspectives are therefore needed to integrate new improvements that will satisfy user needs and LULC monitoring requirements. Key methodological aspects—such as learning from multi-year data, the use of temporal composites versus raw time-series data, the use of a single classification model versus multiple specialized models, and the definition of standardized classification schemes that support the integration of multi-source field data—have been explored and reported in the literature (e.g., ESA WorldCereal project) contributing to the refinement of operational classification workflows.

In light of these advances and monitoring requirements, our study specifically investigated the incorporation of geospatial metadata—such as pixel geographic locations and biogeographical region information—to improve classification performance. In particular, we examined how this information can be effectively integrated into state-of-the-art DL methods and proposed a novel classification framework. This approach explicitly incorporates both fine- and coarse-grained spatial information, allowing DL models to fully exploit the geospatial context of the input data. The results indicate that incorporating geospatial metadata enhances the predictive performance (measured by the F1-score) of land cover classification at the EU scale. Specifically, for a broad 7-class land cover classification (Level-1), the F1-score improves by 1 percentage point. For a more detailed 19-class crop type classification (Level-2), the F1-score increases by 3 percentage points. The most significant improvements are observed when both fine- and coarse-grained spatial information are utilized together. Compared to Random Forest, the improvements are 2 percentage points for the Level-1 classification and 5 percentage points for the Level-2 classification, reaching 80.30\% and 64.01\%, respectively.

In this study, the experiments were conducted using data distributed across the entire EU area; however, a wall-to-wall map was not generated. Ensuring that the models and classification frameworks are scalable and can generalize well across different geographical regions and scales (e.g., local, continental, global) remains a key challenge. Future research should focus on developing models that are computationally efficient and applicable at large scales while maintaining high accuracy and robustness when applied to new and diverse data sets. Additionally, it is crucial to address the specific needs of end-users, such as policymakers, environmental managers, and researchers, to ensure that the developed models and frameworks are practical and beneficial for real-world applications.

\begin{comment}
\section{Data}
\label{sec:data}
\input{data}
\end{comment}

% Author contributions (CRediT)
\section*{CRediT authorship contribution statement}
\textbf{Babak Ghassemi}: Methodology, Software, Writing, and Reviewing
\textbf{Cassio Fraga-Dantas}: Conceptualization, Methodology, Software, Supervision, Writing, and Reviewing
\textbf{Raffaele Gaetano}: Software, Writing, and Reviewing
\textbf{Dino Ienco}: Conceptualization, Methodology, Software, Supervision, Writing, and Reviewing
\textbf{Emma Izquierdo-Verdiguier}: Supervision, Writing, and Reviewing
\textbf{Francesco Vuolo}: Conceptualization, Methodology, Supervision, Writing, and Reviewing 
\textbf{Omid Ghorbanzadeh}: Writing, and Reviewing

% Conflict of interest
\section*{Declaration of competing interest}
The authors declare no competing interests.

% Acknowledgments
\section*{Acknowledgments}
This research was also co-funded by the European Union’s Horizon Europe Research and Innovation Programme under Grant Agreement No 101060423 LAMASUS project.

% Bibliography
\bibliographystyle{elsarticle-harv}
\bibliography{References}

\end{document}